\documentclass[10pt,twocolumn,letterpaper]{article}

\PassOptionsToPackage{usenames,dvipsnames}{xcolor}

\usepackage{iccv}
\usepackage{times}

\usepackage{url}
\usepackage{amsfonts}
\usepackage{amsmath,amssymb}
\usepackage{nicefrac}
\usepackage{microtype}
\usepackage{xcolor}
\usepackage{xspace}
\usepackage{balance}

\usepackage{epsfig}
\usepackage{graphicx}
\usepackage{wrapfig}
\usepackage{subcaption}

\usepackage{multirow}
\usepackage{rotating}
\usepackage{booktabs}
\usepackage{tabularx}
\usepackage{colortbl}

\usepackage{paralist}

\usepackage[pagebackref=true,breaklinks=true,colorlinks,bookmarks=false]{hyperref}

\iccvfinalcopy

\def\iccvPaperID{*****}

\newcolumntype{s}{>{\columncolor[gray]{.85}[.5\tabcolsep]}c}

\newcommand{\xhdr}[1]{\vspace{2pt}\noindent\textbf{#1}}

\newcommand{\tabref}[1]{Tab.~\ref{#1}\xspace}
\newcommand{\figref}[1]{Fig.~\ref{#1}\xspace}
\newcommand{\secref}[1]{Sec.~\ref{#1}\xspace}

\renewcommand{\paragraph}[1]{\noindent\textbf{#1}}

\begin{document}

\title{Navigating to Objects Specified by Images}

\author{
  Jacob Krantz$^1$\thanks{Work done while interning at Meta AI's FAIR Labs.\newline Correspondence: \href{mailto:krantzja@oregonstate.edu}{krantzja@oregonstate.edu}} \qquad
  Theophile Gervet$^2$ \qquad
  Karmesh Yadav$^3$ \qquad
  Austin Wang$^3$
  \\
  Chris Paxton$^3$ \qquad
  Roozbeh Mottaghi$^{3,4}$ \qquad
  Dhruv Batra$^{3,5}$ \qquad
  Jitendra Malik$^{3,6}$
  \\
  Stefan Lee$^1$ \qquad
  Devendra Singh Chaplot$^3$
  \\
  {\normalsize$^1$Oregon State} \quad
  {\normalsize$^2$Carnegie Mellon} \quad
  {\normalsize$^3$Meta AI}  \quad
  {\normalsize$^4$University of Washington}  \quad
  {\normalsize$^5$Georgia Tech}  \quad
  {\normalsize$^6$UC Berkeley}
  \\
  {\normalsize Project Page \& Videos: \href{https://jacobkrantz.github.io/modular_iin}{jacobkrantz.github.io/modular\_iin}}
}

\maketitle
\ificcvfinal\thispagestyle{empty}\fi

\begin{abstract}
   Images are a convenient way to specify which particular object instance an embodied agent should navigate to. Solving this task requires semantic visual reasoning and exploration of unknown environments. We present a system that can perform this task in both simulation and the real world. Our modular method solves sub-tasks of exploration, goal instance re-identification, goal localization, and local navigation. We re-identify the goal instance in egocentric vision using feature-matching and localize the goal instance by projecting matched features to a map. Each sub-task is solved using off-the-shelf components requiring zero fine-tuning. On the HM3D InstanceImageNav benchmark, this system outperforms a baseline end-to-end RL policy 7x and a state-of-the-art ImageNav model 2.3x (56\% \vs 25\% success). We deploy this system to a mobile robot platform and demonstrate effective real-world performance, achieving an 88\% success rate across a home and an office environment.
\end{abstract}

\begin{figure}[ht]
    \centering
    \vspace{20pt}
    \includegraphics[width=0.95\columnwidth]{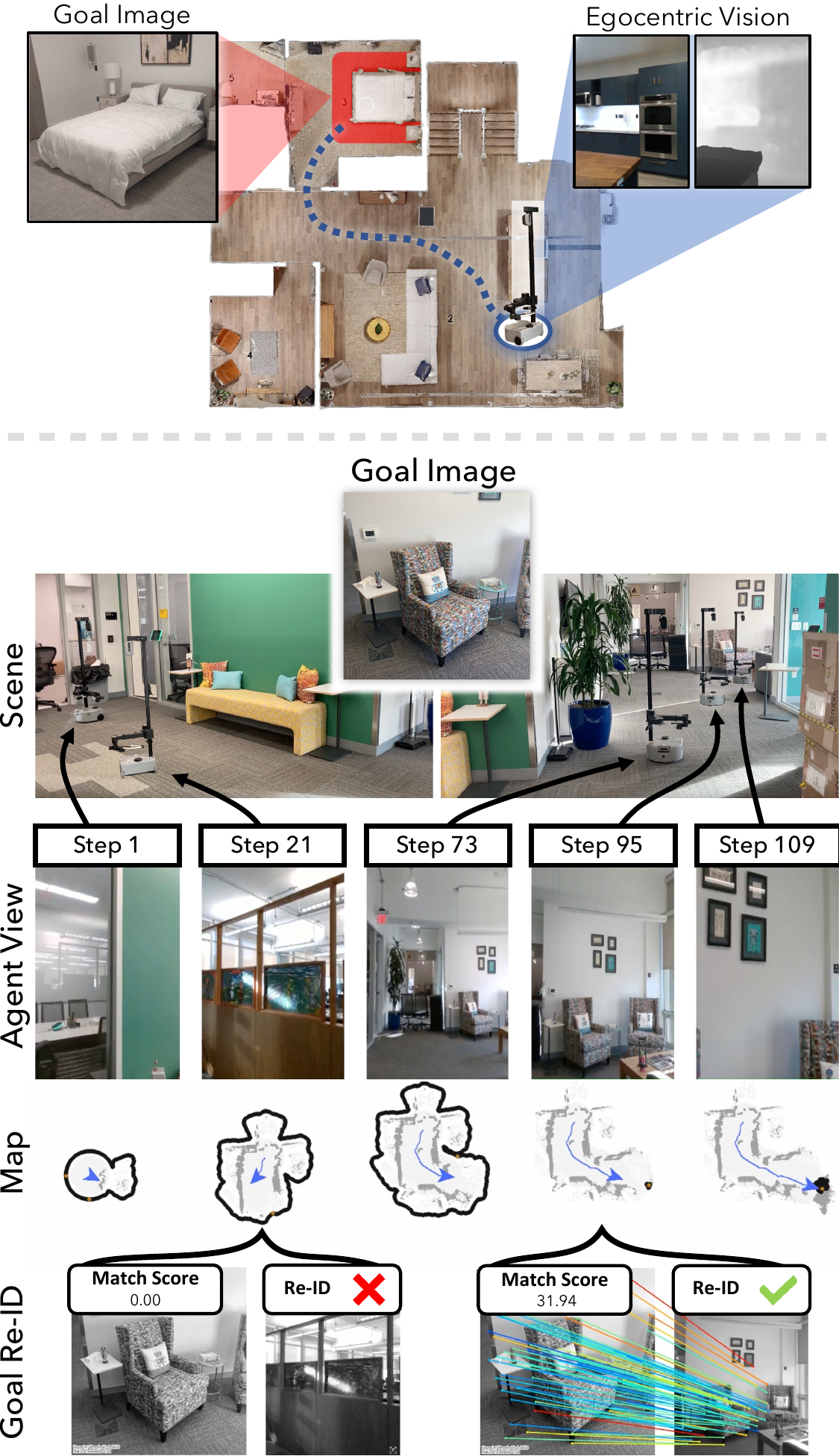}
    \vspace{-5pt}
    \caption{\textbf{Top: InstanceImageNav} tasks an agent with navigating to an object instance described by a goal image.
    \textbf{Bottom: Real-World Deployment.} Our method achieves leading performance in sim and transfers to reality. Here, we find a chair 10m away. Videos are on the \href{https://jacobkrantz.github.io/modular_iin}{project page}.
    }
    \vspace{-30pt}
    \label{fig:teaser}
\end{figure}

\section{Introduction}
\label{sec:intro}

\begin{figure*}[t]
    \centering
    \includegraphics[width=\textwidth]{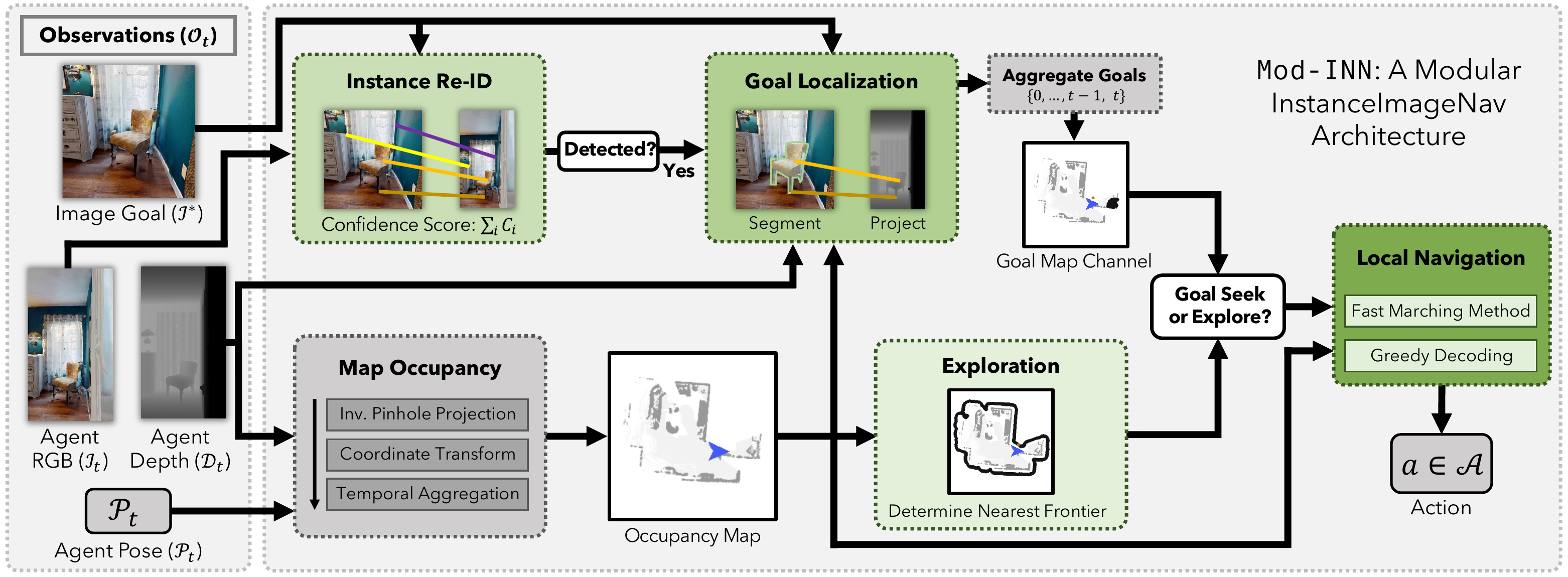}
    \caption{\textbf{Model Overview.} We instantiate \textit{exploration} with frontier-based exploration, \textit{instance re-identification} with feature matching, \textit{goal localization} with masked feature projection, and \textit{local navigation} with analytical planning.
    Sub-task modules are \textcolor{ForestGreen}{green} with supporting components in \textcolor{gray}{gray}.
    }
    \label{fig:model}
\end{figure*}

Imagine you want your robot assistant to check whether you left your laptop on your bed. One way to describe the task is using natural language: “\textit{Check if my laptop is on the bed}”, but what if there are multiple beds in your home? An alternate way is to show an image of your bed to the robot. An image is a convenient way to specify a particular object instance. For example, in \figref{fig:teaser} (Top), the robot starts near the kitchen, is shown an image of a bed, and must navigate to that particular bed. This Image Goal Navigation task requires reasoning over the relation of objects in the scene (\eg, disambiguating between instances of similar appearance) and exploring efficiently to discover where the goal is (\eg, entering bedrooms while searching for the bed).

In this paper, we present a navigation system that can reliably perform this instance-based Image Goal Navigation task in the real-world. Specifically, we propose a modular framework consisting of Exploration, Instance Re-Identification, Goal Localization, and Local Navigation. We instantiate this framework using a simple combination of off-the-shelf components requiring zero fine-tuning. Depicted in \figref{fig:model}, our system uses a frontier-based exploration policy, re-identifies goals with feature matching, localizes goals with projected feature matches, and path plans with an analytical planner. On the challenging Habitat-Matterport3D (HM3D) ~\cite{ramakrishnan2021hm3d} InstanceImageNav benchmark~\cite{krantz2022instance}, we achieve a success rate of $56\%$ \vs $25\%$ for the best baseline. We deploy our system on a mobile robot platform in two real-world environments where it achieves a success rate of $88\%$ (\eg, \figref{fig:teaser} Bottom).

Most prior work tackling Image Goal Navigation (ImageNav)\cite{zhu2017target,chaplot2020neural,hahn2021no,mezghan2022memory,al2022zero,majumdar2022zson,yadav2022offline} assumes that goal images were captured at random poses in the environment and always match the camera parameters of the agent. As argued in Krantz, \etal \cite{krantz2022instance}, this formulation may result in ambiguous image goals (\eg, captures of nondescript walls) and is detached from potential user applications. To overcome these issues, the instance-based ImageNav task (InstanceImageNav) proposed by Krantz, \etal \cite{krantz2022instance} has two key properties: (1) goal images depict an object instance, and (2) goal images are independent of agent embodiment. A baseline end-to-end reinforcement learning policy achieves just 8.3\% success. Our system outperforms this 7-fold. To further compare to prior work, we evaluate a state-of-the-art ImageNav method~\cite{yadav2023ovrl} on InstanceImageNav. While it outperforms the baseline, our system outperforms it 2.3x.

Many prior navigation approaches, such as the two mentioned above, involve training sensors-to-action policies \textit{end-to-end} using reinforcement or imitation learning. While end-to-end methods can be easy to implement and applicable to multiple tasks, they suffer from limitations of high sample complexity~\cite{krantz2022instance}, overfitting~\cite{wu2019bayesian}, and poor sim-to-real transfer~\cite{gervet2022navigating}. Alternatively, navigation can be decomposed into sub-tasks solved with more constrained skills~\cite{chaplotlearning, chaplot2020object,gadre2022clip,gervet2022navigating}. This \textit{modular} paradigm provides benefits of increased sample efficiency and improved real-world execution~\cite{ ramakrishnan2022poni, gervet2022navigating, deitke2022retrospectives}. Our method demonstrates these benefits with efficient sample complexity (zero fine-tuning) and effective real-world performance (\figref{fig:teaser} Bottom), all while achieving top performance on the simulation benchmark.

Altogether, we tackle the important and challenging task of Instance-specific Image Goal Navigation. We propose a modular method that tops the HM3D InstanceImageNav benchmark and outperforms a state-of-the-art ImageNav model. We deploy our system on a mobile robot platform and demonstrate that it can operate in indoor real world environments reliably.

\section{Related Work}
\label{sec:related}

\xhdr{Image Goal Navigation.}
In embodied navigation, targets can be specified via coordinates \cite{anderson2018evaluation} or through various forms of semantic description \cite{batra2020objectnav,zhu2017target,chen2020soundspaces,anderson2018vision,krantz2020beyond}. Image-goal navigation (ImageNav) is one form of the latter; agents navigate in response to a visual description provided by an image. ImageNav is commonly studied in previously-unseen environments \cite{zhu2017target} where goal images are sampled randomly throughout the scene \cite{chaplot2020neural}. While many works study indoor environments, some study outdoor \cite{shah2021ving, shah2021rapid, shah2022viking}. Many approaches to solving ImageNav adopt deep reinforcement learning (DRL) to learn end-to-end policies that map egocentric vision to action \cite{zhu2017target,al2022zero,majumdar2022zson,yadav2022offline}. However, skills relating to visual scene understanding, semantic exploration, and long-term memory tend to be difficult to learn end-to-end. Thus, these methods tend to adopt a combination of careful reward shaping \cite{choi2021image}, pre-training routines \cite{yadav2022offline}, and advanced memory modules \cite{savinovsemi,mezghan2022memory,wu2019bayesian}. In opposition to end-to-end DRL, some approaches carve out sub-tasks that can be learned in a supervised manner such as graph prediction via topological SLAM \cite{chaplot2020neural}, graph-based distance learning \cite{hahn2021no,shah2021rapid}, and camera pose estimation for last-mile navigation \cite{wassermanlast}. In this work, we address the ImageNav task where goal images depict object instances (InstanceImageNav~\cite{krantz2022instance}). We study how far we can drive performance on this benchmark using a purely modular method with no fine-tuning and demonstrate superior performance to a state-of-the-art end-to-end policy.

\xhdr{Modular Methods for Semantic Navigation.}
Classical approaches to navigating previously-unseen environments involve building a geometric map and localizing the agent (SLAM \cite{durrant2006simultaneous}). Modular methods to semantic navigation decompose high-level tasks into components that can either leverage the classical navigation pipeline or be solved with modern vision systems, such as object detectors \cite{he2017mask}. One related task is Object Goal Navigation (ObjectNav~\cite{batra2020objectnav}) in which an agent is given an object category (\eg, \textit{potted plant}) and must navigate to any instance of that category. The ObjectNav task was decomposed by Chaplot \etal \cite{chaplot2020object} to exploration, object detection, and local navigation. Expanding on this, CLIP on Wheels (CoW \cite{gadre2022clip}) employed a decomposition of exploration and object localization to address an open-set object vocabulary. Modular methods also hold promise for transferring systems from simulation to reality (Sim2Real). Gervet \etal \cite{gervet2022navigating} performed Sim2Real transfer of modular and end-to-end ObjectNav systems, finding that modularity avoided the visual Sim2Real gap that degraded the end-to-end policy. In this work, we expand the capability of modular navigation agents to include image-specified navigation targets. We propose a factorization of the problem that can be solved via modular components and demonstrate its effectiveness in both simulation and reality.

\xhdr{Instance Re-Identification.}
Object instance re-identification (OIRe-ID) is the task of determining if a given image depicts the same object depicted in an anchor image. OIRe-ID is commonly operationalized in an image retrieval context \cite{bansal2019re,bansal2021did}. Prior to the deep learning revolution, image retrieval and related visual recognition tasks were based primarily on local feature descriptors~\cite{liu2020deep} such as SIFT~\cite{lowe1999object} or HOG~\cite{dalal2005histograms}. A foundational method of object retrieval involved applying text retrieval methods to these local features, resulting in a bag-of-visual-words model (BoVW~\cite{sivic2003video}). Models relying on local features are limited by the expressivity of feature representation. As such, some modern approaches update this stack with deep networks for description (\eg SuperPoint~\cite{detone2018superpoint}) and matching (\eg SuperGlue \cite{sarlin2020superglue}). Alternatively, re-ranking methods using transformers \cite{tan2021instance} and epipolar-guided transformers \cite{bhalgat2022light} have been proposed in the OIRe-ID space. In this work, we propose solving an embodied OIRe-ID problem where novel instance views are acquired through egomotion. We show that our SuperGlue-based keypoint method not only enables re-identification but also informs localizing the instance.

\section{InstanceImageNav Task Setup}

We follow the task setup proposed by Krantz \etal \cite{krantz2022instance}. The instance-specific image goal navigation task (InstanceImageNav) places an agent at a random pose in an unexplored indoor environment and tasks the agent with navigating to a particular object instance depicted as the primary subject of an RGB image. We match our task setup with the ImageNav track of the 2023 Habitat Navigation Challenge\footnote{\href{https://aihabitat.org/challenge/2023/}{aihabitat.org/challenge/2023}}~\cite{habitatchallenge2023} to enable both clear comparisons and a smooth transfer from simulation to reality.

\xhdr{Observation Space.}
At each time step $t$, the agent’s observation $\mathcal{O}_t$ consists of the RGB goal image $\mathcal{I}^*$, an egocentric RGB image $\mathcal{I}_t$, an egocentric depth image $\mathcal{D}_t$, and the agent’s pose $\mathcal{P}_t = (x, y, \theta)$ relative to the starting pose $\mathcal{P}_0 = (0, 0, 0)$. Collectively, $\mathcal{O}_t = (\mathcal{I}^*, \mathcal{I}_t, \mathcal{D}_t, \mathcal{P}_t)$. The camera capturing $\mathcal{I}^*$ is independent of the camera capturing $\mathcal{I}_t$. Camera parameters for $\mathcal{I}^*$ are sampled to reflect realistic deployments, both extrinsic (height, look-at-angle, distance-to-object) and intrinsic (field of view). See Krantz, \etal \cite{krantz2022instance} for more details. $\mathcal{I}^*$ is a $512 \times 512$ RGB image. The agent's egocentric perception matches the Intel RealSense camera: both $\mathcal{I}_t$ and $\mathcal{D}_t$ are aligned to $640\times360$ with a $42^{\circ}$ HFOV.

\xhdr{Action Space.}
We adopt a discrete action space $\mathcal{A}$ = $\{$\texttt{MOVE-FORWARD}, \texttt{TURN-LEFT}, \texttt{TURN-RIGHT}, \texttt{STOP}$\}$ where forward movement translates the agent by $0.25$m and turn commands rotate the agent by $30^{\circ}$.

\xhdr{Success Criteria.}
An InstanceImageNav episode is successful if the agent issues the \texttt{STOP} action while within $1.0$m Euclidean distance of the object instance depicted by $\mathcal{I}^*$. The target instance must also be oracle-viewable by any combination of turning the agent and looking up or down.

\xhdr{Embodiment.}
Our real-world execution is performed using Stretch by Hello Robot\footnote{\href{https://hello-robot.com/stretch-2}{hello-robot.com/stretch-2}}. We model this embodiment in our simulation experiments with a rigid-body, zero-turn-radius cylinder of height $1.41$m and radius $0.17$m. The forward-facing RGBD camera is mounted at a height of $1.31$m.

\section{Method}
\label{sec:method}

We propose factorizing the InstanceImageNav task into sub-tasks that can be individually addressed. Specifically, we carve out exploration, goal instance re-identification, goal localization, and local navigation to solve InstanceImageNav in aggregate. We describe these sub-tasks as follows.

\xhdr{Exploration.}
Finding an object instance in a previously-unknown environment requires exploration --- both the location of the goal and the map of the environment are unknown. In InstanceImageNav, the goal is described by $\mathcal{I}^*$, where a successful navigation entails observing an $\mathcal{I}_t$ at some time $t$ with a high semantic similarity to $\mathcal{I}^*$. Efficiently visiting locations in the environment that may afford this view, \ie, maximizing \textit{coverage} of the observable space, can lead to a successful navigation.

\xhdr{Goal Instance Re-Identification.}
If an exploration policy results in 100\% coverage of the observable space, then there exists at least one time step $t$ such that the associated image $\mathcal{I}_t$ depicts the goal instance described by $\mathcal{I}^*$. Goal instance re-identification is thus the binary classifier $f_{ID}$ that answers 
\textit{``is the object depicted in $\mathcal{I}^*$ visible in $\mathcal{I}_t$?"}  Concretely,
\begin{align}
    \hat{y_t} = f_{ID}(\mathcal{I}^*, \mathcal{I}_t). 
\end{align}
This task requires leveraging foreground and background to re-identify an object from novel views and is studied extensively in the object instance re-identification (OIRe-ID) literature.

\xhdr{Goal Localization.}
Agents may be far from the goal instance when it is identified, so simply calling \texttt{STOP} is insufficient for success. Thus, it is essential to use the pairing between $\mathcal{I}^*$ and $\mathcal{I}_t$ to localize the goal. Goal localization, $f_{GL}$, maps the paired RGB images and egocentric depth to the position of the goal instance relative to the agent's current pose: 
\begin{align} \label{eq:goal_loc}
    \left ( 
        \mathcal{P}^{(x,y,\cdot)}_G - \mathcal{P}^{(x,y,\theta)}_t
    \right ) = f_{GL}(\mathcal{I}^*, \mathcal{I}_t, \mathcal{D}_t).
\end{align}

\xhdr{Local Navigation.}
Local Navigation is the foundation that enables both exploration and navigation to the goal instance. We consider a local navigation policy $\pi$ that maps a relative polar coordinate goal $(r, \theta)$ to a sequence of actions $\{ a_0, a_1, \dots, a_n \} \in \mathcal{A}$. $\pi$ can be conditioned on a map and/or egocentric vision.

\subsection{Proposed Modules}
\label{sec:method_actual}

We instantiate a system that operationalizes the above factorization: \texttt{Mod-IIN}. Specifically, we propose a purely modular method that can perform InstanceImageNav without \textit{any} re-trained or fine-tuned components. This method is in stark contrast to the prevailing paradigm of learned end-to-end ImageNav policies, yet demonstrates compelling results in simulation and reality. We visualize this model in \figref{fig:model}.

\xhdr{Exploration.}
We adopt a frontier-based exploration (FBE~\cite{yamauchi1997frontier}) policy that operates on a top-down 2D map tracking occupancy, free-space, and frontiers, where frontiers delineate the boundary between explored and unexplored regions. This map is updated each time step using an inverse perspective projection of egocentric depth ($\mathcal{D}_t$) and pose ($\mathcal{P}_t$). FBE greedily selects the nearest frontier in the map to navigate to. Upon reaching that frontier, the process repeats with the selection of the next nearest frontier. This policy enables an efficient expansion of coverage with demonstrated effectiveness in both simulation and reality \cite{gervet2022navigating}.

\xhdr{Goal Instance Re-Identification.}
We employ a keypoint-based re-identification method that performs binary classification conditioned on the goal image and egocentric image. First, we extract the pixel-wise $(x,y)$ location of keypoints $K^* \in \mathbb{R}^{n \times 2}$ in the goal image and their associated vector descriptions $V^* \in \mathbb{R}^{n \times 256}$. We extract these features using SuperPoint~\cite{detone2018superpoint}, a single-pass convolutional neural network trained using homographic adaptation, a self-supervised consistency method. We repeat this for the egocentric image, resulting in $K_t \in \mathbb{R}^{m \times 2}$ and $V_t \in \mathbb{R}^{m \times 256}$. Concretely:
\begin{align}
    (K^*, V^*) &= \text{SuperPoint}(\mathcal{I}^*) \\
    (K_t, V_t) &= \text{SuperPoint}(\mathcal{I}_t)
\end{align}
We then compute correspondences between $(K^*, V^*)$ and $(K_t, V_t)$, resulting in a matched subset $(\hat{K}^* \in \mathbb{R}^{w \times 2}, \hat{K}_t \in \mathbb{R}^{w \times 2})$ such that the $i^{\text{th}}$ keypoint of $\hat{K}^*$ corresponds to the $i^{\text{th}}$ keypoint of $\hat{K}_t$ with $w \leq \text{min}(n, m)$. Each keypoint pair has a match confidence score $0 \leq C \in \mathbb{R}^{w} \leq 1$. We use SuperGlue~\cite{sarlin2020superglue}, a graph neural network (GNN) that optimizes a partial match assignment via optimal transport:
\begin{align}
    (\hat{K}^*, \hat{K}_t), C &= \text{SuperGlue}((K^*, V^*), (K_t, V_t))
\end{align}
We can then turn this output into a binary classifier by thresholding the sum of the confidence scores:
\begin{align}
    \text{sum}(C) \geq \tau
\end{align}
where a positive truth value indicates re-identification of the goal instance and $\tau$ is a chosen threshold. This classifier is evaluated on $(\mathcal{I}^*, \mathcal{I}_t)$ at each step $t$. For both feature extraction (SuperPoint) and feature matching (SuperGlue), we use off-the-shelf models with zero downstream fine-tuning.

\begin{figure*}[t]
    \centering
    \begin{subfigure}{0.33\textwidth}
        \centering
        \includegraphics[width=0.65\textwidth]{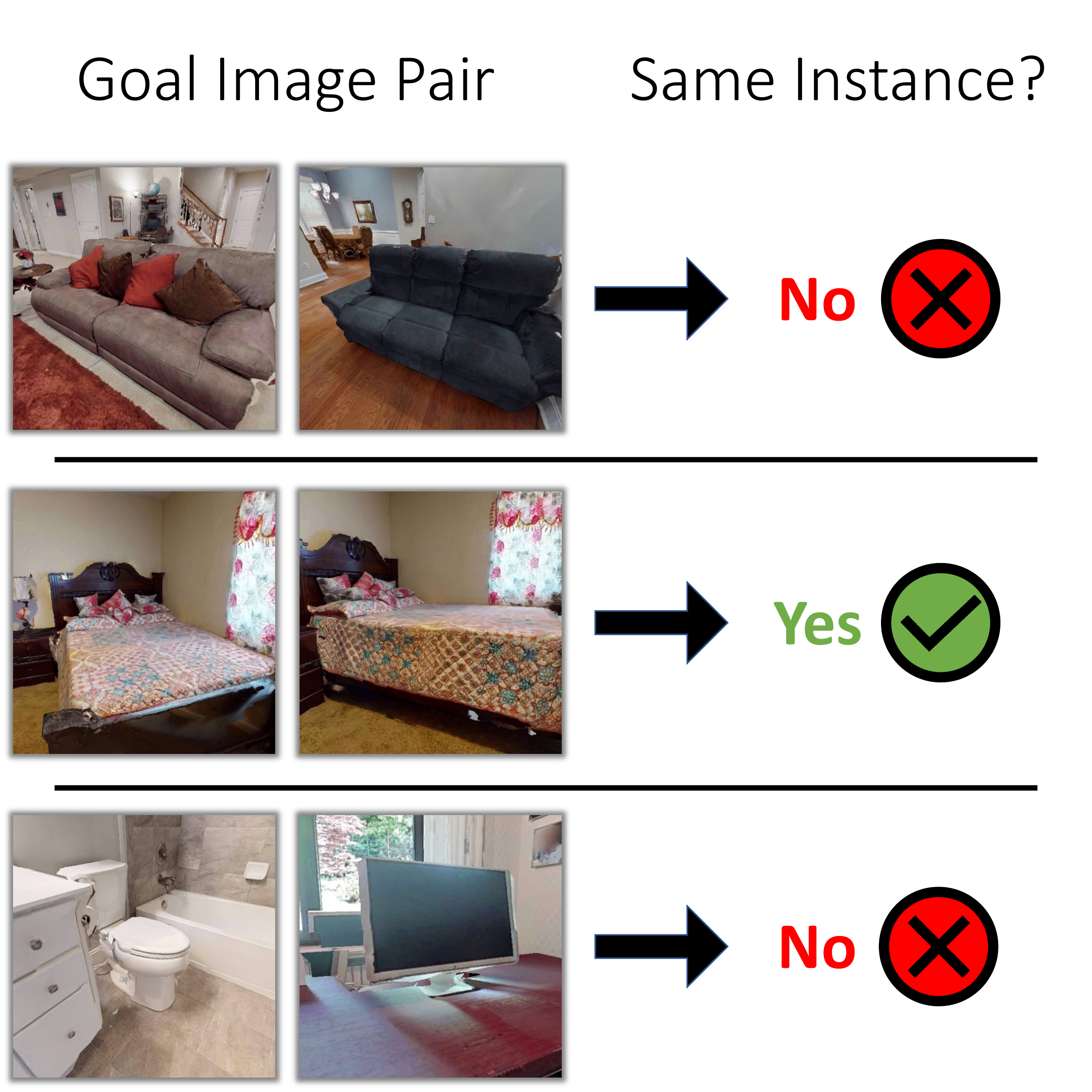}
        \vspace*{5mm}
        \caption{Classification Dataset}
        \label{fig:ds}
    \end{subfigure}
    \begin{subfigure}{0.33\textwidth}
        \centering
        \includegraphics[width=\textwidth]{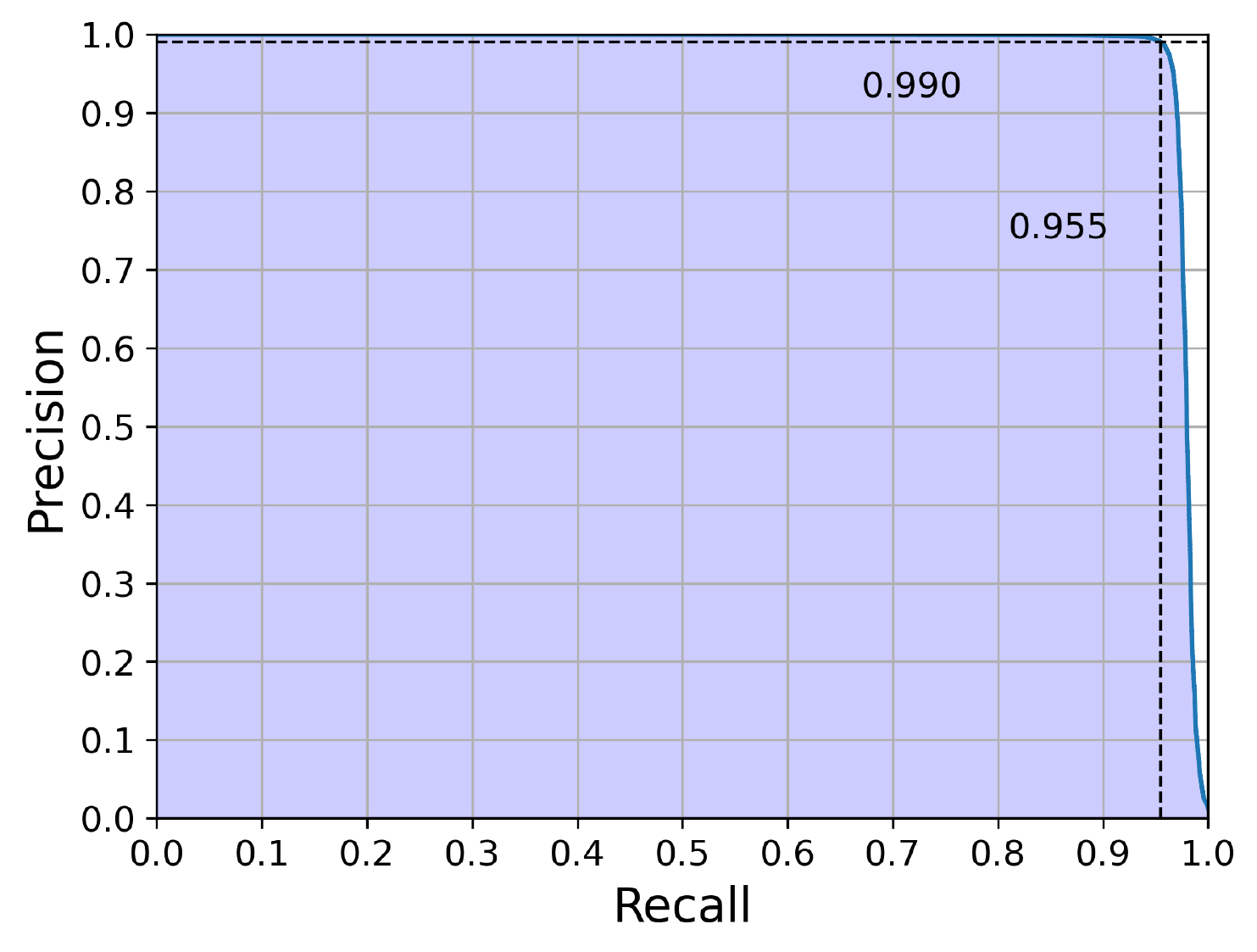}
        \caption{Precision-Recall Curve}
        \label{fig:pr_curve}
    \end{subfigure}
    \begin{subfigure}{0.33\textwidth}
        \centering
        \includegraphics[width=\textwidth]{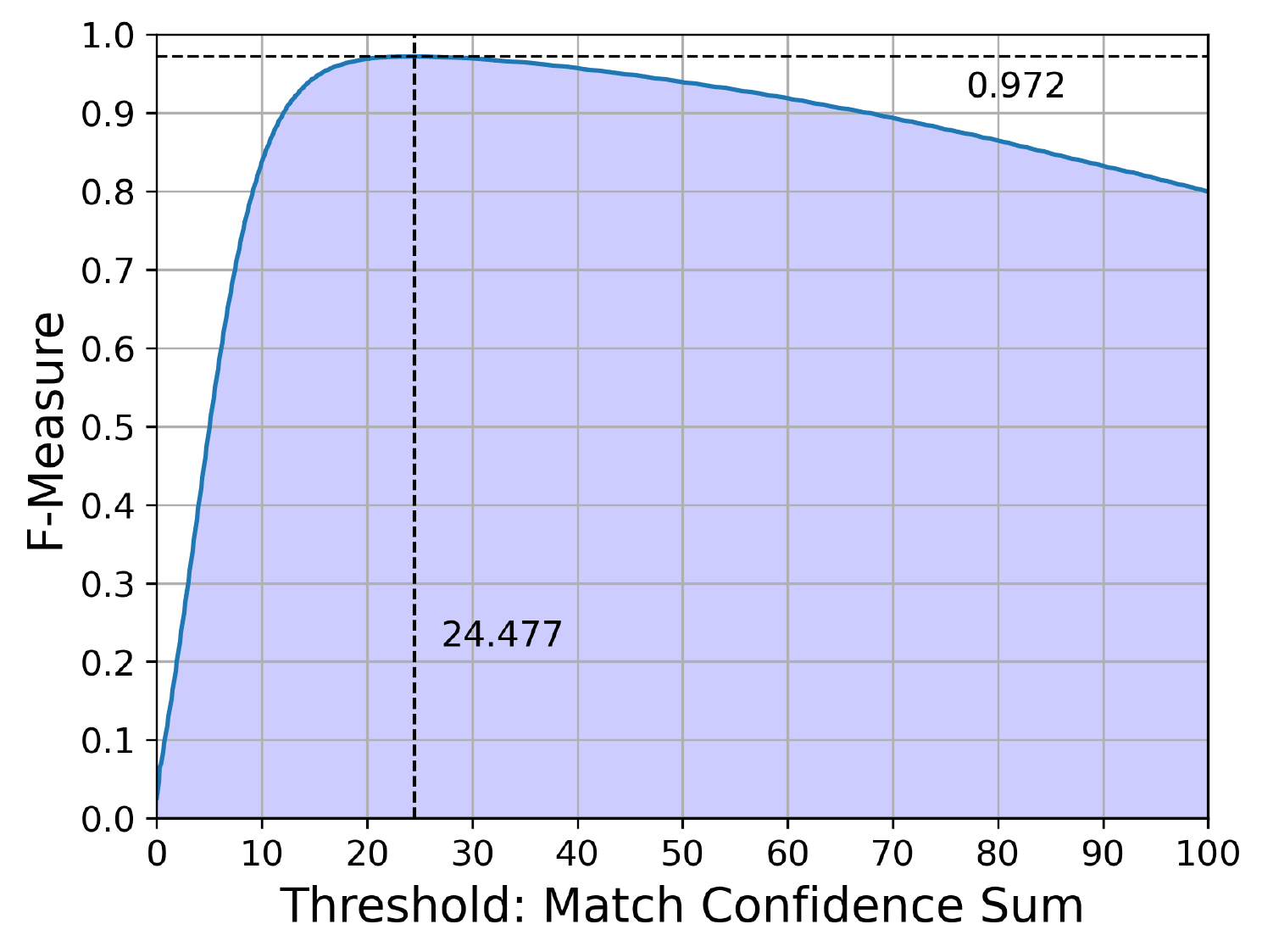}
        \caption{F-Measure Curve}
        \label{fig:f_curve}
    \end{subfigure}\hfill
    \caption{To determine an instance re-identification (Re-ID) threshold $\tau$, we collect a dataset of goal image pairs (a). Our Re-ID method computes the sum of feature matching scores for each image pair which can then be used to compute a PR curve~(b) and an f-measure curve~(c). We select the $\tau$ with maximal f-measure ($\tau=24.5$).}
    \label{fig:thresh_curves}
\end{figure*}

\xhdr{Determining a Detection Threshold $\tau$.}
If $\tau$ is too low, egocentric images that do not observe the goal will be incorrectly passed on to goal localization (too many false positives). If $\tau$ is too high, steps during exploration that observe the goal will be missed (too many false negatives). To strike a balance between these error modes, we collect a dataset of pairs of goal images and pick the threshold that maximizes our classifier's F-measure. This method is as follows.

We sample a set of object instances that are represented in the training split of the InstanceImageNav episode dataset. Each object has between 1 and 50 goal images that depict it. We extract $3\%$ of the object instances and their associated goal images, resulting in $121$ objects and $2270$ goal images (an average of $18.8$ images per object). We use this data to construct a pairwise dataset where the input consists of two images $(\mathcal{I}_a, \mathcal{I}_b)$ and the output $y$ consists of whether those images observe the same object: $ ((\mathcal{I}_a, \mathcal{I}_b), y) \in D$. We then run our classifier on all image pairs to produce a set of confidence scores. This enables us to compute precision-recall (PR) and F-measure curves for our classifier (\figref{fig:thresh_curves}). In the end, we select the threshold that maximizes F-measure (the harmonic mean of precision and recall). We verify that this produces an empirically optimal threshold in Sup. \ref{supp:optimal}.

\xhdr{Goal Localization.}
Upon observing the goal instance, we need to localize it in the world (Eq. \ref{eq:goal_loc}). From the previous re-identification step, we have correspondences $(\hat{K}^*, \hat{K}_t)$ between the goal image and the egocentric image. Using the depth map $\mathcal{D}_t$ aligned to our egocentric image, we can project these points into the world via an inverse perspective projection. However, not all matched keypoints lie on the goal instance we seek to navigate to; visual features associated with the background and adjacent objects can cause the agent to stop at the wrong location.
The question becomes, \textit{``which matched keypoints should we project as goal targets?"}.

We perform instance segmentation of the goal image to determine a mask of the goal instance --- keypoints inside the mask can be mapped to the egocentric frame and then be projected. We perform instance segmentation using an off-the-shelf pre-trained network (Detic~\cite{zhou2022detecting}). We select Detic as it is a high-performance detector that allows for an open vocabulary by encoding concepts through CLIP~\cite{radford2021learning}. In using Detic, our method is not constrained to fixed object categories and can be extended to image goals depicting arbitrary objects. To determine a goal instance mask, we select the mask containing the image center point with highest confidence.

All keypoints in $\hat{K}^*$ that lie within this mask are mapped to their corresponding keypoint in $\hat{K}_t$ and projected to the world. The point cloud of goal points is coalesced along the height dimension and voxelized into a 2D map channel. This channel is then concatenated with the map constructed during exploration and treated as a navigation target.

\xhdr{Local Navigation.}
The local navigator must navigate to frontier points (exploration) and goal points (goal localization). Both can be addressed as follows; given a partial occupancy map and a set of goal points, plan a path in the agent's action space $\mathcal{A}$ that conveys the agent to the nearest reachable point. We solve this using an incremental path planner based on the fast marching method (FMM~\cite{sethian1996fast}) as proposed by \cite{chaplotlearning} and used in subsequent works \cite{chaplot2020object,hahn2021no,krantz2022sim}.

\section{Experiments}
\label{sec:experiments}

We evaluate our model in simulation and reality. In simulation, we compare to prior art and alternate sub-task modules (\secref{sec:exp-results}). We analyze failure modes (\secref{sec:exp-failures}) and discuss a qualitative example (\secref{sec:exp-qual}). Finally, we demonstrate successful Sim2Real transfer (\secref{sec:exp-real}).

\subsection{Experimental Setup}

\xhdr{Simulation.} 
Our simulated experiments follow the task definition and dataset proposed for the ImageNav track of the 2023 Habitat Navigation Challenge~\cite{krantz2022instance}. We build and evaluate our agent on top of the Habitat Simulator~\cite{savva2019habitat, szot2021habitat}. The episode dataset follows the generation procedure proposed by Krantz, \etal \cite{krantz2022instance} for InstanceImageNav with the additional constraint that multi-floor navigation is not required of any episode. Scenes supporting this dataset come from the Habitat-Matterport3D Dataset~\cite{ramakrishnan2021hm3d} with semantic annotations (HM3D-SEM~\cite{yadav2022habitat}). The 216 scenes in HM3D-SEM are split Train/Val/Test on 145/36/35 and InstanceImageNav episodes are split Train/Val/Test-Standard/Test-Challenge on 7,056K/1K/1K/1K. The object instances depicted by the goal images have a category belonging to one of the following: $\{$ \textit{chair, couch, plant, bed, toilet, television} $\}$. We use the Val split which is composed of 795 unique object instances.

\xhdr{Metrics.}
We evaluate models on success and efficiency. Particularly, we report success rate (\texttt{SR}), success weighted by inverse path length (\texttt{SPL}), navigation error (\texttt{NE}), and, in some analyses, maximum steps taken (\texttt{Max-ST}). Success is true if upon calling \texttt{STOP}, the agent is within $1.0$m Euclidean distance of the goal and the object is oracle-visible by turning or looking up/down. \texttt{SPL} is an efficiency measure defined in \cite{anderson2018evaluation} and modified for goal viewpoints in \cite{batra2020objectnav}. \texttt{NE} is the geodesic distance between the agent’s stopping location and the nearest goal viewpoint. \texttt{Max-ST} is the percentage of episodes that use the full step budget ($1000$).

\xhdr{Baselines.} We compare our agent to the following models:
\begin{itemize}
    \item \texttt{IIN RL Baseline}: The InstanceImageNav (IIN) baseline model \cite{krantz2022instance}  is an end-to-end sensors-to-action recurrent policy consisting of unimodal encoders for the goal image, egocentric image, egocentric depth, and pose. This model was trained from scratch using reinforcement learning for 3.5 billion steps of experience using proximal policy optimization (PPO~\cite{schulman2017proximal}) with variable experience rollout (VER~\cite{wijmansver}).
    \item \texttt{OVRL-v2}: Offline Visual Representation learning V2 (\texttt{OVRL-v2}~\cite{yadav2023ovrl}) is a model-free, end-to-end semantic navigation policy that employs a ViT+LSTM architecture and self-supervised visual pre-training. At the time of writing, \texttt{OVRL-v2} achieves state-of-the-art performance on a prior ImageNav benchmark and near state-of-the-art on an ObjectNav benchmark. We evaluate the pre-trained model zero-shot on InstanceImageNav (\texttt{OVRL-v2-ImageNav}). To make a fair comparison, we then take the pre-trained model and fine-tune it for InstanceImageNav (\texttt{OVRL-v2-IIN}) following the same training routine and parameters used in \cite{yadav2023ovrl} for their ImageNav experiments.
\end{itemize}

\setlength{\tabcolsep}{.3em}
\begin{table}[t]
    \setlength{\aboverulesep}{0pt}
    \setlength{\belowrulesep}{0pt}
    \renewcommand{\arraystretch}{1.15}
    \begin{center}
	\resizebox{0.68\columnwidth}{!}{
		\begin{tabular}{cl csc}
			\toprule
            & 
			& \multicolumn{3}{c}{\scriptsize\textbf{Validation}}
            \\
			\cmidrule{3-5}
			\scriptsize \shortstack{\#} &
			{\scriptsize Model}
			& \scriptsize\textbf{\texttt{NE}}~$\downarrow$
			& \scriptsize\textbf{\texttt{SR}}~$\uparrow$
			& \scriptsize\textbf{\texttt{SPL}}~$\uparrow$
			\\
			\midrule
			\scriptsize \texttt{1}
                    & \texttt{IIN RL Baseline}
                    &  6.3  & 0.083  &  0.035
                    \\
			\scriptsize \texttt{2}
                    & \texttt{OVRL-v2-ImageNav}
                    &  6.9  & 0.006  &  0.002
                    \\
			\scriptsize \texttt{3}
                    & \texttt{OVRL-v2-IIN}
                    &  5.0  & 0.248  & 0.118
                    \\
			\scriptsize \texttt{4}
                    & \texttt{Mod-IIN} (Ours)
                    &  \textbf{3.1}  & \textbf{0.561}  &  \textbf{0.233}
                    \\
			\bottomrule
		\end{tabular}}
	\end{center}
	\caption{We compare our Modular InstanceImageNav method (\texttt{Mod-IIN}) against prior methods on InstanceImageNav. \texttt{Mod-IIN} outperforms the baseline model (row 1) with a $6.8$x increase in \texttt{SR} and outperforms a state-of-the-art ImageNav model (OVRL-v2, row 3) by $2.3$x.}
	\label{tab:prior_art}
\end{table}

\xhdr{Ablations: Goal Instance Re-Identification.} We compare our goal instance re-identification method against baseline approaches. We consider:
\begin{itemize}
	\item \textit{Keypoint-Conf:} This is our primary method as described in \secref{sec:method_actual}. In summary, keypoints are extracted from both goal and egocentric images and matched. This is converted to binary classification by thresholding the sum of correspondence confidence scores.
	\item \textit{Keypoint-Match:}  This method is the same as above but with a different classification strategy; the number of matched keypoint pairs are thresholded instead of the confidence sum. This allows us to test if match confidence affords additional discriminative value.
	\item \textit{ResNet:} This method ablates keypoints entirely by encoding both the goal and the egocentric images with a ResNet-50~\cite{he2016deep} pre-trained on ImageNet. We compute and threshold the cosine similarity between the resulting feature vectors such that a cosine similarity above $\tau$ implies re-identification of the goal instance.
	\item \textit{CLIP:} This method is the same as \texttt{ResNet} but encodes images using a model trained contrastively: CLIP~\cite{radford2021learning}.
	\item \textit{Oracle:} This method provides an upper bound to the instance re-identification sub-task by querying the simulator for an oracle instance mask in the agent's egocentric frame. A positive detection is made if any pixel in the mask matches the ID of the goal object.
\end{itemize}
All methods above (except for \textit{Oracle}) threshold some scalar value to perform detection. Each threshold is determined via the maximal F-measure method presented in \secref{sec:method_actual}. We include PR and F-measure curves for each in Sup. \ref{supp:thresholds}.

\xhdr{Ablations: Goal Localization.} We compare our goal localization method against baseline approaches. We consider:
\begin{itemize}
	\item \textit{Detic-Projected:} This is our primary method as described in \secref{sec:method_actual}. In summary, matched goal image keypoints that lie within the goal instance mask (computed by Detic) are projected to a goal map channel.
	\item \textit{Crop-Projected:} This method ablates instance segmentation via Detic. Matched keypoints that lie within a static central crop of the goal image are projected. We provide further details of this method in Sup. \ref{supp:crop}.
	\item \textit{Detic-ObjectNav:} This method ablates keypoints. Detic infers the object category of the goal instance. Upon positive Re-ID, Detic identifies all pixels in the egocentric image that belong to the inferred category. These pixels are projected and the agent navigates to the closest object of that class. This method takes the semantic segmentation form of modular ObjectNav~\cite{chaplot2020object,gervet2022navigating}.
	\item \textit{Oracle:} This method provides an upper bound to the goal localization sub-task. Upon positive detection, all egocentric pixels that observe the goal object instance (according to an oracle instance mask) are projected.
\end{itemize}

\subsection{Simulation Results}
\label{sec:exp-results}

\setlength{\tabcolsep}{.3em}
\begin{table}[t]
    \setlength{\aboverulesep}{0pt}
    \setlength{\belowrulesep}{0pt}
    \renewcommand{\arraystretch}{1.15}
    \begin{center}
	\resizebox{1.00\columnwidth}{!}{
		\begin{tabular}{cll ccsc}
			\toprule
            & \multicolumn{2}{c}{\scriptsize\textbf{Model Variation}}
			& \multicolumn{4}{c}{\scriptsize\textbf{Validation}}
            \\
			\cmidrule{4-7}
			\scriptsize \shortstack{\#} &
			Instance Re-ID
                & Goal Localization
			& \scriptsize\textbf{\texttt{NE}}~$\downarrow$
			& \scriptsize\textbf{\texttt{Max-ST}}~$\downarrow$
			& \scriptsize\textbf{\texttt{SR}}~$\uparrow$
			& \scriptsize\textbf{\texttt{SPL}}~$\uparrow$
			\\
			\midrule
			\scriptsize \texttt{1}
                    & Keypoint-Conf
                    & Detic-Projected
                    &  \textbf{3.096}  & \textbf{0.310} & \textbf{0.561}  &  \textbf{0.233}
                    \\
			\scriptsize \texttt{2}
                    & Keypoint-Match
                    & Detic-Projected
                    &  3.261  & 0.339 & 0.539  &  0.221
                    \\
			\scriptsize \texttt{3}
                    & Keypoint-Conf
                    & Crop-Projected
                    &  3.128  & 0.301 & 0.523  &  0.224
                    \\
			\scriptsize \texttt{4}
                    & Keypoint-Conf
                    & Detic-ObjectNav
                    &  3.235  & 0.327 & 0.488  &  0.205
                    \\
			\scriptsize \texttt{5}
                    & ResNet
                    & Detic-ObjectNav
                    &  6.264  & 0.857 & 0.138  &  0.044
                    \\
			\scriptsize \texttt{6}
                    & CLIP
                    & Detic-ObjectNav
                    &  6.570  & 0.917 & 0.097  &  0.029
                    \\
			\midrule
			\scriptsize \texttt{7}
                    & Oracle
                    & Oracle
                    &  1.300  & 0.089 & 0.845  &  0.453
                    \\
			\scriptsize \texttt{8}
                    & Oracle
                    & Detic-Projected
                    &  2.551  & 0.116 & 0.498  &  0.250
                    \\
			\scriptsize \texttt{9}
                    & Keypoint-Conf
                    & Oracle
                    &  2.914  & 0.348 & 0.647  &  0.266
                    \\
			\bottomrule
		\end{tabular}}
	\end{center}
	\caption{Our Modular InstanceImageNav method (\texttt{Mod-IIN}: row 1) with variations to instance re-identification and goal localization. }
	\label{tab:ablations}
\end{table}

We center our results discussion on key observations.

\xhdr{\texttt{Mod-IIN} Outperforms Prior Art.}
Our method outperforms the baseline InstanceImageNav method (\tabref{tab:prior_art} row 1 \vs 4) with a nearly 7-fold increase in success (0.561 \texttt{SR} \vs 0.083 \texttt{SR}). \texttt{Mod-IIN} requires zero fine-tuning, while the end-to-end \texttt{IIN RL Baseline} was trained for 3.5 billion steps of experience distributed across 64 GPUs.

Evaluating OVRL-v2 zero-shot on InstanceImageNav (\tabref{tab:prior_art} row 2) results in very low performance ($0.006$ \texttt{SR}). Contributing factors include a different scene dataset (Gibson \vs HM3D), a change in embodiment (LocoBot \vs Stretch), and a different goal destination (image source \vs image subject). OVRL-v2 trained for InstanceImageNav (\texttt{OVRL-v2-IIN}) performs better with a $0.248$ \texttt{SR} (\tabref{tab:prior_art} row 3), yet despite its visual pre-training, demonstrates significant overfitting to the Train split ($0.850$ \texttt{SR}). Our method performs 2.3x better in Val (\tabref{tab:prior_art} row 4 \vs 3).

\xhdr{Keypoint Confidence Is More Discriminative Than Match Count.}
Thresholding the sum of match confidence scores (\textit{Keypoint-Conf}) leads to a downstream improvement of 0.022 \texttt{SR} over thresholding the count of matched keypoints (\textit{Keypoint-Match}) (\tabref{tab:ablations} row 1 \vs 2). This result is supported by comparing the maximal F-measure of the two classifiers on the pair-wise goal image dataset; \textit{Keypoint-Match} produces 0.972 while \textit{Keypoint-Conf} produces 0.962.

\xhdr{Goal Localization Benefits From Instance Segmentation.}
Replacing the instance mask (\textit{Detic-Projected}) with a central crop (\textit{Crop-Projected}) reduces success by 0.038 (\tabref{tab:ablations} row 1 \vs 3). Beyond worse performance, the central crop method exploits a size bias in the dataset that may not generalize.

\xhdr{Keypoint-Based Localization Outperforms Class-Based Localization.}
Both methods of matched keypoint projection, \textit{Detic-Projected} and \textit{Crop-Projected}, outperform class-based localization (\textit{Detic-ObjectNav}) (\tabref{tab:ablations} rows 1,3 \vs 4). Semantic segmentation methods common for goal localization in ObjectNav cannot be trivially applied to InstanceImageNav. This validates that navigating to object instances in indoor environments fundamentally requires the ability to disambiguate between object instances of the same class.

\xhdr{Off-The-Shelf Image Encoders Fail to Re-ID Instances.}
Both \textit{ResNet} and \textit{CLIP} fail to discriminate views of object instances. This is evident in \tabref{tab:ablations} rows 5\&6 where \texttt{Max-ST} is 86\% and 92\%, leading to success rates of 14\% and 10\%, respectively. We found reducing the \textit{ResNet} detection threshold failed to improve performance.

\begin{figure}[t]
    \centering
    \includegraphics[width=\columnwidth]{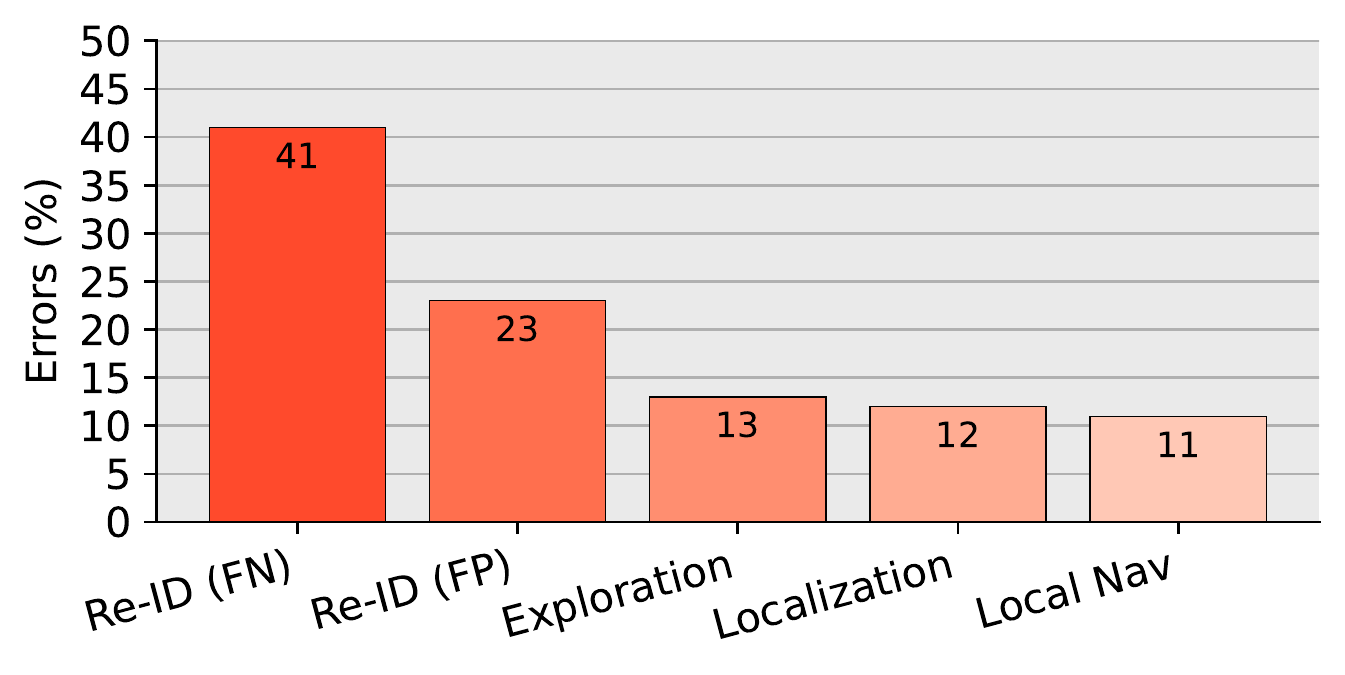}
    \caption{A distribution of \texttt{Mod-IIN} failure modes.}
    \label{fig:failures}
\end{figure}

\subsection{Failure Analysis.}
\label{sec:exp-failures}

In \tabref{tab:ablations} row 7, we evaluate our agent with perfect instance Re-ID and goal localization, meaning that any failures can be attributed to exploration via FBE with local navigation. The resulting performance of 0.845 \texttt{SR} acts as an upper bound when addressing instance Re-ID and goal localization within our system. In row 8, we use oracle instance Re-ID with predicted localization. Performance does not improve over our complete model. We suspect this to be caused by the shared reliance on keypoint matching between our detection and localization systems; assuming detection with low match confidence leads to poor localization. Finally, row 9 demonstrates a 0.086 \texttt{SR} gap between our goal localization method and perfect goal localization (row 9 \vs 1).

We present qualitative analysis of \texttt{Mod-IIN} failure modes in \figref{fig:failures}. For a random subset of 100 failed episodes, we label the cause of failure as one of the following:

\xhdr{Instance Re-ID: False Negative ($41\%$).}
Most commonly, the agent observes and fails to detect the goal. Improved Re-ID methods may mitigate this, but not all novel instance views are equally challenging; Re-ID would be simplified if exploration produces views more similar to the goal image.

\xhdr{Instance Re-ID: False Positive ($23\%$).}
Incorrect detections often come from visual features correctly matched to the goal image background. This failure mode could be mitigated with additional background \vs foreground reasoning.

\xhdr{Exploration Error ($13\%$).}
Exploration fails to produce a view of the goal instance within the allotted time budget.

\xhdr{Localization Error ($12\%$).}
Goal localization errors result from projecting points not belonging to the goal instance, caused either by incorrect correspondences or goal masking.

\xhdr{Local Navigation Error ($11\%$).}
Local navigation occasionally fails to plan a path to the goal.

\begin{figure}[t]
    \centering
    \includegraphics[width=\columnwidth]{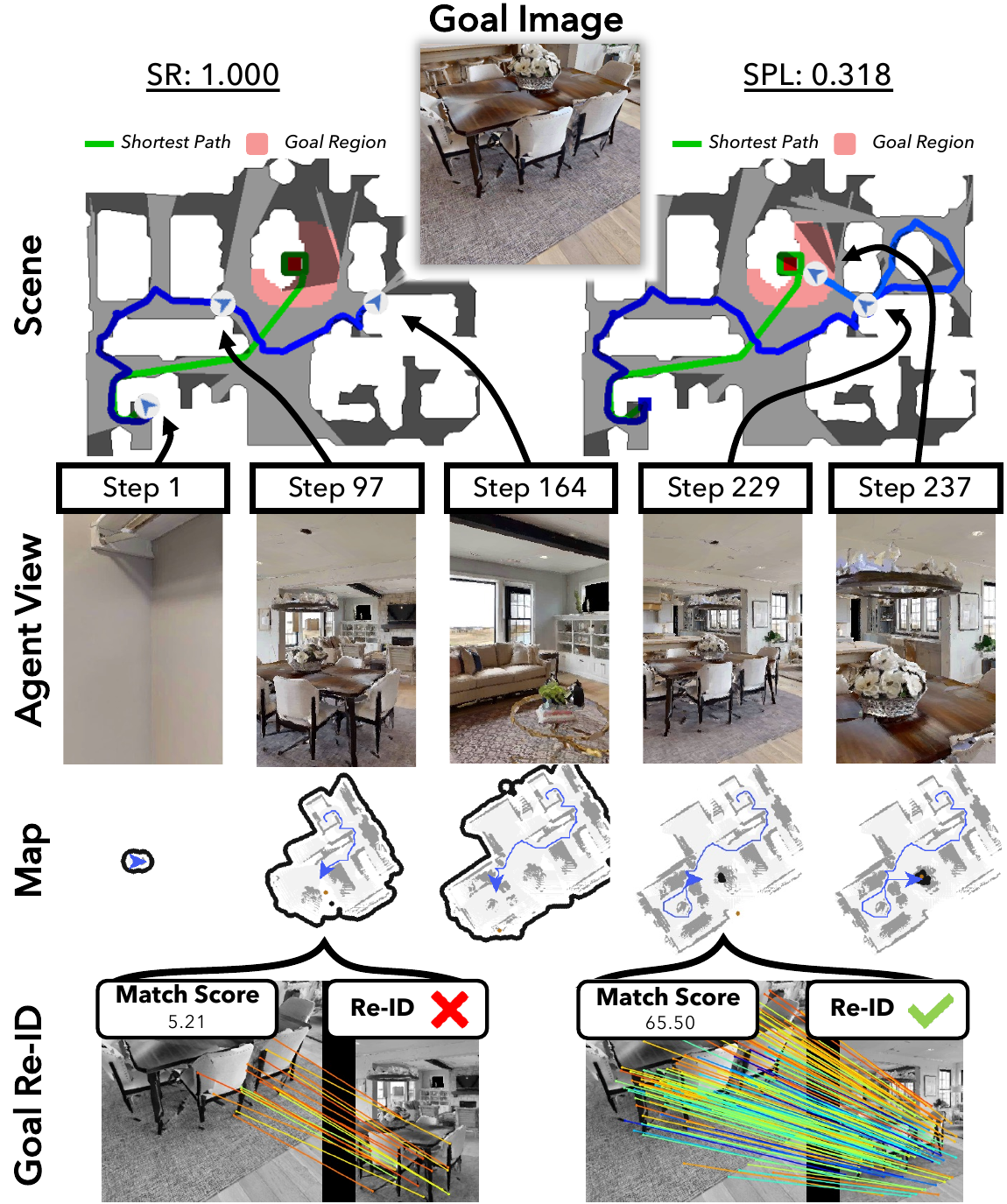}
    \caption{Qualitative example of our \texttt{Mod-IIN} agent performing the InstanceImageNav task in the Habitat Simulator.}
    \label{fig:qualitative2}
\end{figure}

\subsection{Qualitative Example}
\label{sec:exp-qual}

We present a qualitative example of our agent performing an episode in simulation in \figref{fig:qualitative2}. During Steps 1, 97, and 164, the agent is exploring the environment. A detection is made at Step 229, and keypoints are projected as a goal target. Upon reaching the goal at Step 237 (a dining chair), the agent calls \texttt{STOP}. This episode demonstrates a long exploration horizon with a challenging goal instance; multiple identical objects exist in the scene, yet the agent navigated to the correct one. This is due to our instance Re-ID method matching all keypoints in the frame and our localization method only projecting those depicting the goal. We provide videos of our agent in simulation on our \href{https://jacobkrantz.github.io/modular_iin}{project page}.

\subsection{Real-World Deployment}
\label{sec:exp-real}

\begin{figure}[t]
    \centering
    \includegraphics[width=0.9\columnwidth]{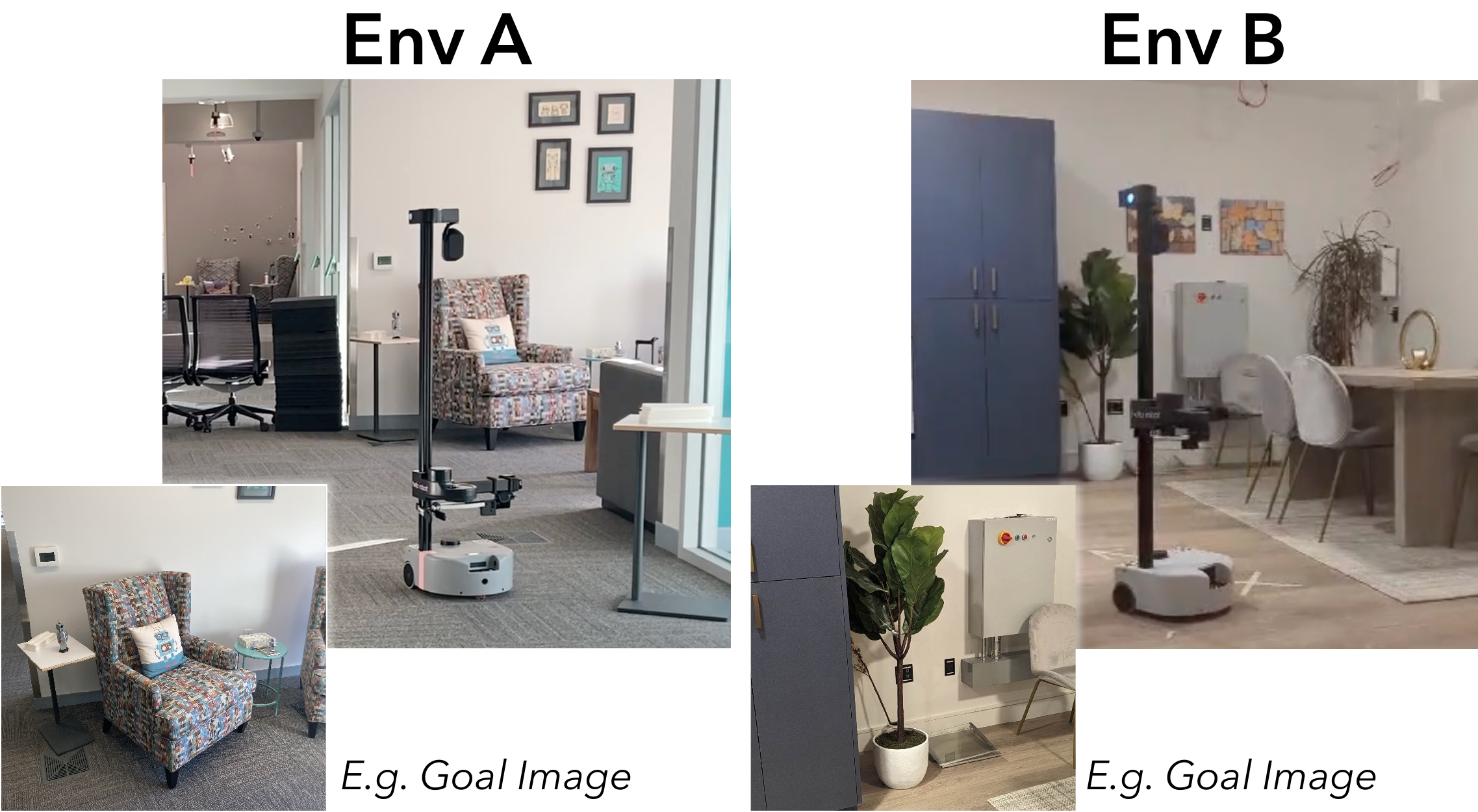}
    \caption{We deploy \texttt{Mod-INN} to a Hello Robot Stretch and evaluate in both an office (Env A) and an apartment (Env B). Our system achieves an 88\% success rate across 8 episodes.}
    \label{fig:real_world}
\end{figure}

We evaluate our method in the real world by deploying to a Hello Robot Stretch~\cite{kemp2022design}. We evaluate in two different environments for a total of eight episodes using five unique image goals (\figref{fig:real_world}). Environment A (Env A) is a furnished office space with a hallway and lounge. Multiple potted plants, couches, chairs and other confounding objects exist in the scene. We experiment with image goals depicting a couch, chair, and potted plant. Environment B (Env B) as depicted in \figref{fig:teaser} is a furnished apartment with a kitchen, living area, bedroom, bathroom, and office. We experiment with image goals depicting a bed and a potted plant. Image goals are captured using a mobile phone. All episodes start without viewing the goal such that exploration is required. The shortest-path geodesic distance of episodes range from 3-10m. Altogether, our agent is successful in $7/8=88\%$ of episodes. Videos of each episode are on our \href{https://jacobkrantz.github.io/modular_iin}{project page}.

\section{Discussion}
\label{sec:discussion}

We decompose InstanceImageNav into exploration, goal re-identification, goal localization, and local navigation. We craft a system within this framework that performs InstanceImageNav with zero fine-tuning. Our experiments show that this system outperforms existing state-of-the-art end-to-end learned policies and transfers to real-world execution.

\xhdr{Limitations and Impact.}
One limitation of our system is that detection and localization are memory-less; performing sequential tasks in persistent environments would benefit from grounding goal images in compressed memory. Our method can serve as a strong and robust baseline for evaluating trained navigation policies and our framework can serve as a catalyst for developing modular semantic navigators.

{
\small
\xhdr{Acknowledgements}
The Oregon State effort is supported in part by the DARPA Machine Common Sense program. The views and conclusions contained herein are those of the authors and should not be interpreted as representing the official policies or endorsements, either expressed or implied, of the US Government or any sponsor.
}

{\small
\bibliographystyle{ieee_fullname}
\bibliography{egbib}

\begin{thebibliography}{10}\itemsep=-1pt

\bibitem{al2022zero}
Ziad Al-Halah, Santhosh~Kumar Ramakrishnan, and Kristen Grauman.
\newblock Zero experience required: Plug \& play modular transfer learning for
  semantic visual navigation.
\newblock In {\em Computer Vision and Pattern Recognition (CVPR)}, 2022.

\bibitem{anderson2018evaluation}
Peter Anderson, Angel Chang, Devendra~Singh Chaplot, Alexey Dosovitskiy,
  Saurabh Gupta, Vladlen Koltun, Jana Kosecka, Jitendra Malik, Roozbeh
  Mottaghi, Manolis Savva, et~al.
\newblock On evaluation of embodied navigation agents.
\newblock {\em arXiv preprint arXiv:1807.06757}, 2018.

\bibitem{anderson2018vision}
Peter Anderson, Qi Wu, Damien Teney, Jake Bruce, Mark Johnson, Niko
  S{\"u}nderhauf, Ian Reid, Stephen Gould, and Anton Van Den~Hengel.
\newblock Vision-and-language navigation: Interpreting visually-grounded
  navigation instructions in real environments.
\newblock In {\em Computer Vision and Pattern Recognition (CVPR)}, 2018.

\bibitem{bansal2021did}
Vaibhav Bansal, Gian~Luca Foresti, and Niki Martinel.
\newblock Where did i see it? object instance re-identification with attention.
\newblock In {\em International Conference on Computer Vision (ICCV)}, 2021.

\bibitem{bansal2019re}
Vaibhav Bansal, Stuart James, and Alessio Del~Bue.
\newblock re-obj: Jointly learning the foreground and background for object
  instance re-identification.
\newblock In {\em International Conference on Image Analysis and Processing
  (ICIAP)}, 2019.

\bibitem{batra2020objectnav}
Dhruv Batra, Aaron Gokaslan, Aniruddha Kembhavi, Oleksandr Maksymets, Roozbeh
  Mottaghi, Manolis Savva, Alexander Toshev, and Erik Wijmans.
\newblock Objectnav revisited: On evaluation of embodied agents navigating to
  objects.
\newblock {\em arXiv preprint arXiv:2006.13171}, 2020.

\bibitem{bhalgat2022light}
Yash Bhalgat, Joao~F Henriques, and Andrew Zisserman.
\newblock A light touch approach to teaching transformers multi-view geometry.
\newblock {\em arXiv preprint arXiv:2211.15107}, 2022.

\bibitem{chaplotlearning}
Devendra~Singh Chaplot, Dhiraj Gandhi, Saurabh Gupta, Abhinav Gupta, and Ruslan
  Salakhutdinov.
\newblock Learning to explore using active neural slam.
\newblock In {\em International Conference on Learning Representations (ICLR)},
  2020.

\bibitem{chaplot2020object}
Devendra~Singh Chaplot, Dhiraj~Prakashchand Gandhi, Abhinav Gupta, and Russ~R
  Salakhutdinov.
\newblock Object goal navigation using goal-oriented semantic exploration.
\newblock In {\em Neural Information Processing Systems (NeurIPS)}, 2020.

\bibitem{chaplot2020neural}
Devendra~Singh Chaplot, Ruslan Salakhutdinov, Abhinav Gupta, and Saurabh Gupta.
\newblock Neural topological slam for visual navigation.
\newblock In {\em Computer Vision and Pattern Recognition (CVPR)}, 2020.

\bibitem{chen2020soundspaces}
Changan Chen, Unnat Jain, Carl Schissler, Sebastia Vicenc~Amengual Gari, Ziad
  Al-Halah, Vamsi~Krishna Ithapu, Philip Robinson, and Kristen Grauman.
\newblock Soundspaces: Audio-visual navigation in 3d environments.
\newblock In {\em European Conference on Computer Vision (ECCV)}, 2020.

\bibitem{choi2021image}
Yunho Choi and Songhwai Oh.
\newblock Image-goal navigation via keypoint-based reinforcement learning.
\newblock In {\em International Conference on Ubiquitous Robots (UR)}, 2021.

\bibitem{dalal2005histograms}
Navneet Dalal and Bill Triggs.
\newblock Histograms of oriented gradients for human detection.
\newblock In {\em Computer Vision and Pattern Recognition (CVPR)}, 2005.

\bibitem{deitke2022retrospectives}
Matt Deitke, Dhruv Batra, Yonatan Bisk, Tommaso Campari, Angel~X Chang,
  Devendra~Singh Chaplot, Changan Chen, Claudia~P{\'e}rez D'Arpino, Kiana
  Ehsani, Ali Farhadi, et~al.
\newblock Retrospectives on the embodied ai workshop.
\newblock {\em arXiv preprint arXiv:2210.06849}, 2022.

\bibitem{detone2018superpoint}
Daniel DeTone, Tomasz Malisiewicz, and Andrew Rabinovich.
\newblock Superpoint: Self-supervised interest point detection and description.
\newblock In {\em Computer Vision and Pattern Recognition workshops (CVPRW)},
  2018.

\bibitem{durrant2006simultaneous}
Hugh Durrant-Whyte and Tim Bailey.
\newblock Simultaneous localization and mapping: part i.
\newblock {\em IEEE robotics \& automation magazine}, 2006.

\bibitem{gadre2022clip}
Samir~Yitzhak Gadre, Mitchell Wortsman, Gabriel Ilharco, Ludwig Schmidt, and
  Shuran Song.
\newblock Clip on wheels: Zero-shot object navigation as object localization
  and exploration.
\newblock {\em arXiv preprint arXiv:2203.10421}, 2022.

\bibitem{gervet2022navigating}
Theophile Gervet, Soumith Chintala, Dhruv Batra, Jitendra Malik, and
  Devendra~Singh Chaplot.
\newblock Navigating to objects in the real world.
\newblock {\em arXiv preprint arXiv:2212.00922}, 2022.

\bibitem{hahn2021no}
Meera Hahn, Devendra~Singh Chaplot, Shubham Tulsiani, Mustafa Mukadam, James~M
  Rehg, and Abhinav Gupta.
\newblock No rl, no simulation: Learning to navigate without navigating.
\newblock In {\em Neural Information Processing Systems (NeurIPS}, 2021.

\bibitem{he2017mask}
Kaiming He, Georgia Gkioxari, Piotr Doll{\'a}r, and Ross Girshick.
\newblock Mask r-cnn.
\newblock In {\em International Conference on Computer Vision (ICCV)}, 2017.

\bibitem{he2016deep}
Kaiming He, Xiangyu Zhang, Shaoqing Ren, and Jian Sun.
\newblock Deep residual learning for image recognition.
\newblock In {\em Computer Vision and Pattern Recognition (CVPR)}, 2016.

\bibitem{kemp2022design}
Charles~C Kemp, Aaron Edsinger, Henry~M Clever, and Blaine Matulevich.
\newblock The design of stretch: A compact, lightweight mobile manipulator for
  indoor human environments.
\newblock In {\em International Conference on Robotics and Automation (ICRA)},
  2022.

\bibitem{krantz2022sim}
Jacob Krantz and Stefan Lee.
\newblock Sim-2-sim transfer for vision-and-language navigation in continuous
  environments.
\newblock In {\em European Conference on Computer Vision (ECCV)}, 2022.

\bibitem{krantz2022instance}
Jacob Krantz, Stefan Lee, Jitendra Malik, Dhruv Batra, and Devendra~Singh
  Chaplot.
\newblock Instance-specific image goal navigation: Training embodied agents to
  find object instances.
\newblock {\em arXiv preprint arXiv:2211.15876}, 2022.

\bibitem{krantz2020beyond}
Jacob Krantz, Erik Wijmans, Arjun Majumdar, Dhruv Batra, and Stefan Lee.
\newblock Beyond the nav-graph: Vision-and-language navigation in continuous
  environments.
\newblock In {\em European Conference on Computer Vision (ECCV)}, 2020.

\bibitem{liu2020deep}
Li Liu, Wanli Ouyang, Xiaogang Wang, Paul Fieguth, Jie Chen, Xinwang Liu, and
  Matti Pietik{\"a}inen.
\newblock Deep learning for generic object detection: A survey.
\newblock {\em International journal of computer vision}, 2020.

\bibitem{lowe1999object}
David~G Lowe.
\newblock Object recognition from local scale-invariant features.
\newblock In {\em International Conference on Computer Vision (ICCV)}, 1999.

\bibitem{majumdar2022zson}
Arjun Majumdar, Gunjan Aggarwal, Bhavika Devnani, Judy Hoffman, and Dhruv
  Batra.
\newblock Zson: Zero-shot object-goal navigation using multimodal goal
  embeddings.
\newblock {\em arXiv preprint arXiv:2206.12403}, 2022.

\bibitem{mezghan2022memory}
Lina Mezghan, Sainbayar Sukhbaatar, Thibaut Lavril, Oleksandr Maksymets, Dhruv
  Batra, Piotr Bojanowski, and Karteek Alahari.
\newblock Memory-augmented reinforcement learning for image-goal navigation.
\newblock In {\em International Conference on Intelligent Robots and Systems
  (IROS)}, 2022.

\bibitem{radford2021learning}
Alec Radford, Jong~Wook Kim, Chris Hallacy, Aditya Ramesh, Gabriel Goh,
  Sandhini Agarwal, Girish Sastry, Amanda Askell, Pamela Mishkin, Jack Clark,
  et~al.
\newblock Learning transferable visual models from natural language
  supervision.
\newblock In {\em International Conference on Machine Learning (ICML)}, 2021.

\bibitem{ramakrishnan2022poni}
Santhosh~K. Ramakrishnan, Devendra~Singh Chaplot, Ziad Al-Halah, Jitendra
  Malik, and Kristen Grauman.
\newblock Poni: Potential functions for objectgoal navigation with
  interaction-free learning.
\newblock In {\em Computer Vision and Pattern Recognition (CVPR)}, 2022.

\bibitem{ramakrishnan2021hm3d}
Santhosh~Kumar Ramakrishnan, Aaron Gokaslan, Erik Wijmans, Oleksandr Maksymets,
  Alexander Clegg, John~M Turner, Eric Undersander, Wojciech Galuba, Andrew
  Westbury, Angel~X Chang, Manolis Savva, Yili Zhao, and Dhruv Batra.
\newblock Habitat-matterport 3d dataset ({HM}3d): 1000 large-scale 3d
  environments for embodied {AI}.
\newblock In {\em Neural Information Processing Systems (NeurIPS) Datasets and
  Benchmarks Track}, 2021.

\bibitem{sarlin2020superglue}
Paul-Edouard Sarlin, Daniel DeTone, Tomasz Malisiewicz, and Andrew Rabinovich.
\newblock Superglue: Learning feature matching with graph neural networks.
\newblock In {\em Computer Vision and Pattern Recognition (CVPR)}, 2020.

\bibitem{savinovsemi}
Nikolay Savinov, Alexey Dosovitskiy, and Vladlen Koltun.
\newblock Semi-parametric topological memory for navigation.
\newblock In {\em International Conference on Learning Representations (ICLR)},
  2018.

\bibitem{savva2019habitat}
Manolis Savva, Abhishek Kadian, Oleksandr Maksymets, Yili Zhao, Erik Wijmans,
  Bhavana Jain, Julian Straub, Jia Liu, Vladlen Koltun, Jitendra Malik, et~al.
\newblock Habitat: A platform for embodied ai research.
\newblock In {\em International Conference on Computer Vision (ICCV)}, 2019.

\bibitem{schulman2017proximal}
John Schulman, Filip Wolski, Prafulla Dhariwal, Alec Radford, and Oleg Klimov.
\newblock Proximal policy optimization algorithms.
\newblock {\em arXiv preprint arXiv:1707.06347}, 2017.

\bibitem{sethian1996fast}
James~A Sethian.
\newblock A fast marching level set method for monotonically advancing fronts.
\newblock {\em Proceedings of the National Academy of Sciences (PNAS)}, 1996.

\bibitem{shah2021ving}
Dhruv Shah, Benjamin Eysenbach, Gregory Kahn, Nicholas Rhinehart, and Sergey
  Levine.
\newblock Ving: Learning open-world navigation with visual goals.
\newblock In {\em International Conference on Robotics and Automation (ICRA)},
  2021.

\bibitem{shah2021rapid}
Dhruv Shah, Benjamin Eysenbach, Nicholas Rhinehart, and Sergey Levine.
\newblock {Rapid Exploration for Open-World Navigation with Latent Goal
  Models}.
\newblock In {\em Conference on Robot Learning (CoRL)}, 2021.

\bibitem{shah2022viking}
Dhruv Shah and Sergey Levine.
\newblock {ViKiNG: Vision-Based Kilometer-Scale Navigation with Geographic
  Hints}.
\newblock In {\em Robotics: Science and Systems (RSS)}, 2022.

\bibitem{sivic2003video}
Josef Sivic and Andrew Zisserman.
\newblock Video google: A text retrieval approach to object matching in videos.
\newblock In {\em International Conference on Computer Vision (ICCV)}, 2003.

\bibitem{szot2021habitat}
Andrew Szot, Alex Clegg, Eric Undersander, Erik Wijmans, Yili Zhao, John
  Turner, Noah Maestre, Mustafa Mukadam, Devendra Chaplot, Oleksandr Maksymets,
  Aaron Gokaslan, Vladimir Vondrus, Sameer Dharur, Franziska Meier, Wojciech
  Galuba, Angel Chang, Zsolt Kira, Vladlen Koltun, Jitendra Malik, Manolis
  Savva, and Dhruv Batra.
\newblock Habitat 2.0: Training home assistants to rearrange their habitat.
\newblock In {\em Neural Information Processing Systems (NeurIPS)}, 2021.

\bibitem{tan2021instance}
Fuwen Tan, Jiangbo Yuan, and Vicente Ordonez.
\newblock Instance-level image retrieval using reranking transformers.
\newblock In {\em International Conference on Computer Vision (ICCV)}, 2021.

\bibitem{wassermanlast}
Justin Wasserman, Karmesh Yadav, Girish Chowdhary, Abhinav Gupta, and Unnat
  Jain.
\newblock Last-mile embodied visual navigation.
\newblock In {\em Conference on Robot Learning (CoRL)}, 2022.

\bibitem{wijmansver}
Erik Wijmans, Irfan Essa, and Dhruv Batra.
\newblock Ver: Scaling on-policy rl leads to the emergence of navigation in
  embodied rearrangement.
\newblock In {\em Neural Information Processing Systems (NeurIPS)}, 2022.

\bibitem{wu2019bayesian}
Yi Wu, Yuxin Wu, Aviv Tamar, Stuart Russell, Georgia Gkioxari, and Yuandong
  Tian.
\newblock Bayesian relational memory for semantic visual navigation.
\newblock In {\em International Conference on Computer Vision (ICCV)}, 2019.

\bibitem{habitatchallenge2023}
Karmesh Yadav, Jacob Krantz, Ram Ramrakhya, Santhosh~Kumar Ramakrishnan, Jimmy
  Yang, Austin Wang, John Turner, Aaron Gokaslan, Oleksandr Maksymets, Angel~X
  Chang, Manolis Savva, Devendra~Singh Chaplot, Alexander Clegg, and Dhruv
  Batra.
\newblock Habitat challenge 2023.
\newblock \url{https://aihabitat.org/challenge/2023/}, 2023.

\bibitem{yadav2023ovrl}
Karmesh Yadav, Arjun Majumdar, Ram Ramrakhya, Naoki Yokoyama, Alexei Baevski,
  Zsolt Kira, Oleksandr Maksymets, and Dhruv Batra.
\newblock Ovrl-v2: A simple state-of-art baseline for imagenav and objectnav.
\newblock {\em arXiv preprint arXiv:2303.07798}, 2023.

\bibitem{yadav2022offline}
Karmesh Yadav, Ram Ramrakhya, Arjun Majumdar, Vincent-Pierre Berges, Sachit
  Kuhar, Dhruv Batra, Alexei Baevski, and Oleksandr Maksymets.
\newblock Offline visual representation learning for embodied navigation.
\newblock {\em arXiv preprint arXiv:2204.13226}, 2022.

\bibitem{yadav2022habitat}
Karmesh Yadav, Ram Ramrakhya, Santhosh~Kumar Ramakrishnan, Theo Gervet, John
  Turner, Aaron Gokaslan, Noah Maestre, Angel~Xuan Chang, Dhruv Batra, Manolis
  Savva, et~al.
\newblock Habitat-matterport 3d semantics dataset.
\newblock {\em arXiv preprint arXiv:2210.05633}, 2022.

\bibitem{yamauchi1997frontier}
Brian Yamauchi.
\newblock A frontier-based approach for autonomous exploration.
\newblock In {\em International Symposium on Computational Intelligence in
  Robotics and Automation (CIRA)}, 1997.

\bibitem{zhou2022detecting}
Xingyi Zhou, Rohit Girdhar, Armand Joulin, Philipp Kr{\"a}henb{\"u}hl, and
  Ishan Misra.
\newblock Detecting twenty-thousand classes using image-level supervision.
\newblock In {\em European Conference on Computer Vision (ECCV)}, 2022.

\bibitem{zhu2017target}
Yuke Zhu, Roozbeh Mottaghi, Eric Kolve, Joseph~J Lim, Abhinav Gupta, Li
  Fei-Fei, and Ali Farhadi.
\newblock Target-driven visual navigation in indoor scenes using deep
  reinforcement learning.
\newblock In {\em International Conference on Robotics and Automation (ICRA)}.
  IEEE, 2017.

\end{thebibliography}
}

\clearpage

\appendix

\section*{Supplementary Material}

\section{Verifying Threshold Optimality}
\label{supp:optimal}

\begin{figure}[t]
    \centering
    \includegraphics[width=0.9\columnwidth]{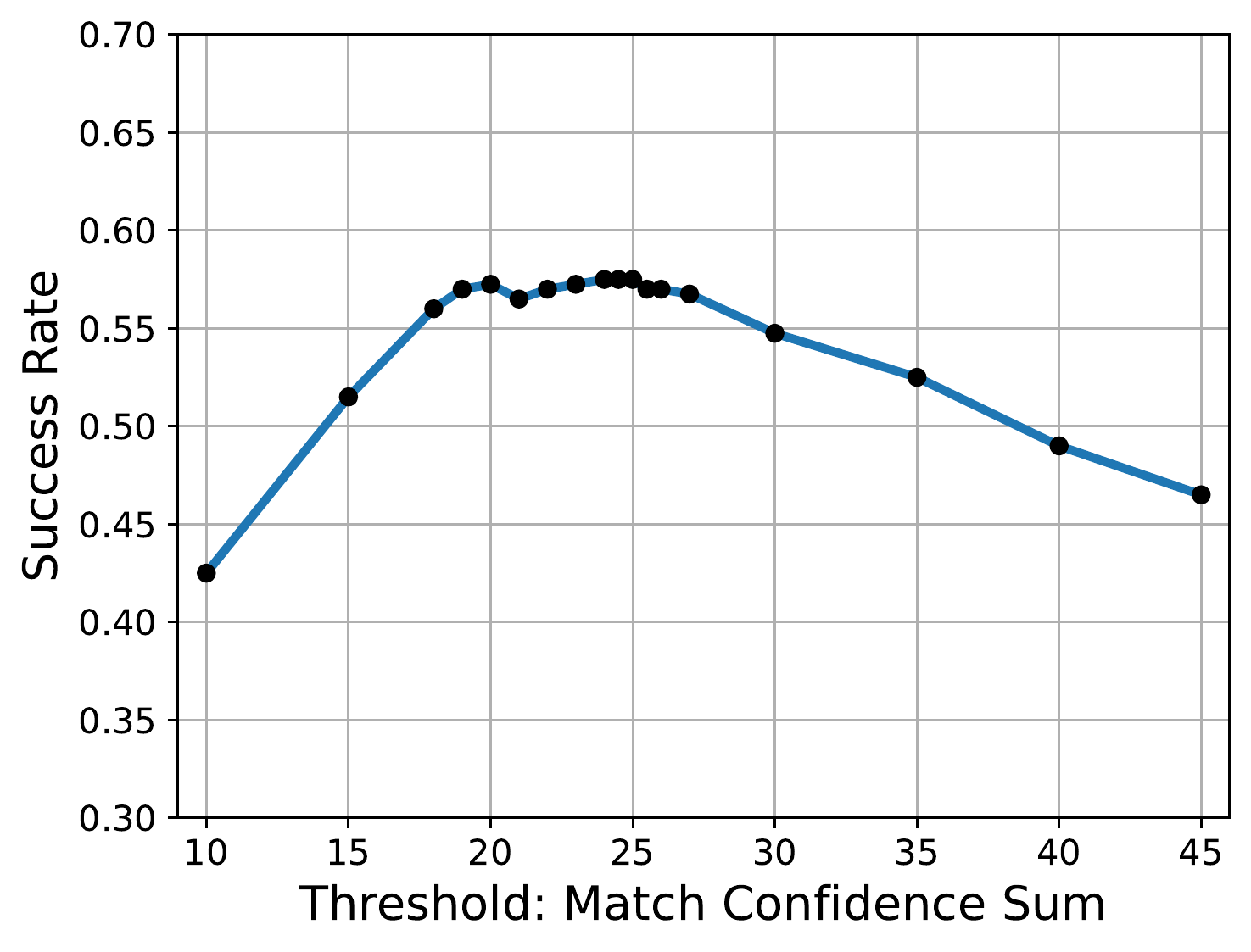}
    \caption{Empirical performance of instance Re-ID thresholds on a subset of the Train split.}
    \label{fig:threshold_search}
\end{figure}

We evaluate the effectiveness of our threshold-determining method for instance re-identification (\secref{sec:method_actual}). We extract a subset of 400 episodes belonging to the Train split and ensure scene and object diversity by sampling from 20 different scenes, 40 object instances per scene, and 1 episode per object. We then perform a search for the best detection threshold by running our complete \texttt{Mod-IIN} system with varying detection thresholds. In \figref{fig:threshold_search}, we see the results of this search with success rate plotted against threshold value. The threshold determined via our maximal F-measure method coincides directly with the optimal threshold in downstream performance.

\section{Re-ID Thresholds of Alternate Methods}
\label{supp:thresholds}

\begin{figure*}[t]
    \centering
    \includegraphics[width=\textwidth]{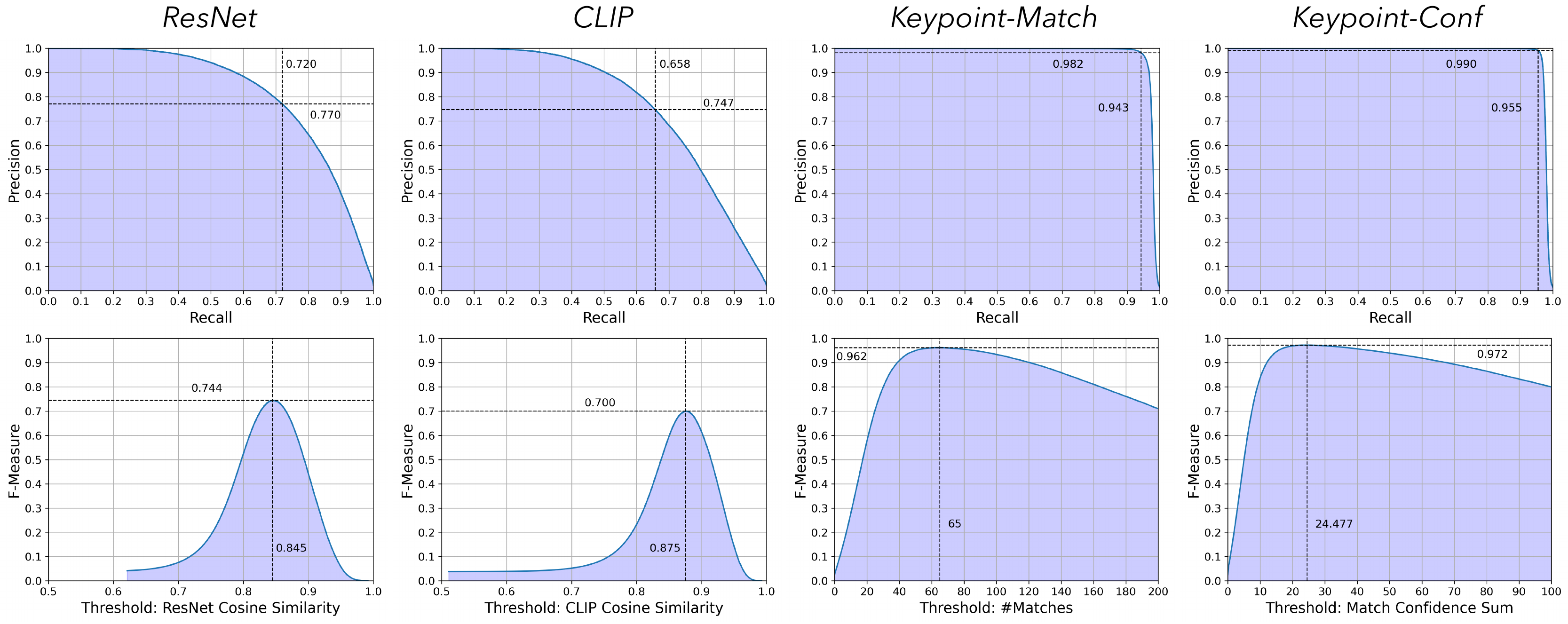}
    \caption{Precision-recall curves (top) and F-measure curves (bottom) for each goal instance Re-ID method.}
    \label{fig:thresholds_all}
\end{figure*}

In \figref{fig:thresholds_all}, we show precision-recall curves and F-measure curves for each goal instance re-identification method we experiment with in the main paper: \textit{ResNet}, \textit{CLIP}, \textit{Keypoint-Match}, and \textit{Keypoint-Conf}. Both keypoint-based methods demonstrate significantly higher PR AUC and maximal F-measure scores than \textit{ResNet} and \textit{CLIP}. While maximal F-measure is very close between \textit{Keypoint-Match} and \textit{Keypoint-Conf} (0.962 \vs 0.972), this difference leads to a downstream InstanceImageNav performance difference of 0.022 \texttt{SR} (\tabref{tab:ablations}). Notably, the rank order of maximal F-measure aligns with the rank order of downstream performance.

\section{\textit{Crop-Projected} Goal Localization Details}
\label{supp:crop}

\begin{figure}[t]
    \centering
    \includegraphics[width=\columnwidth]{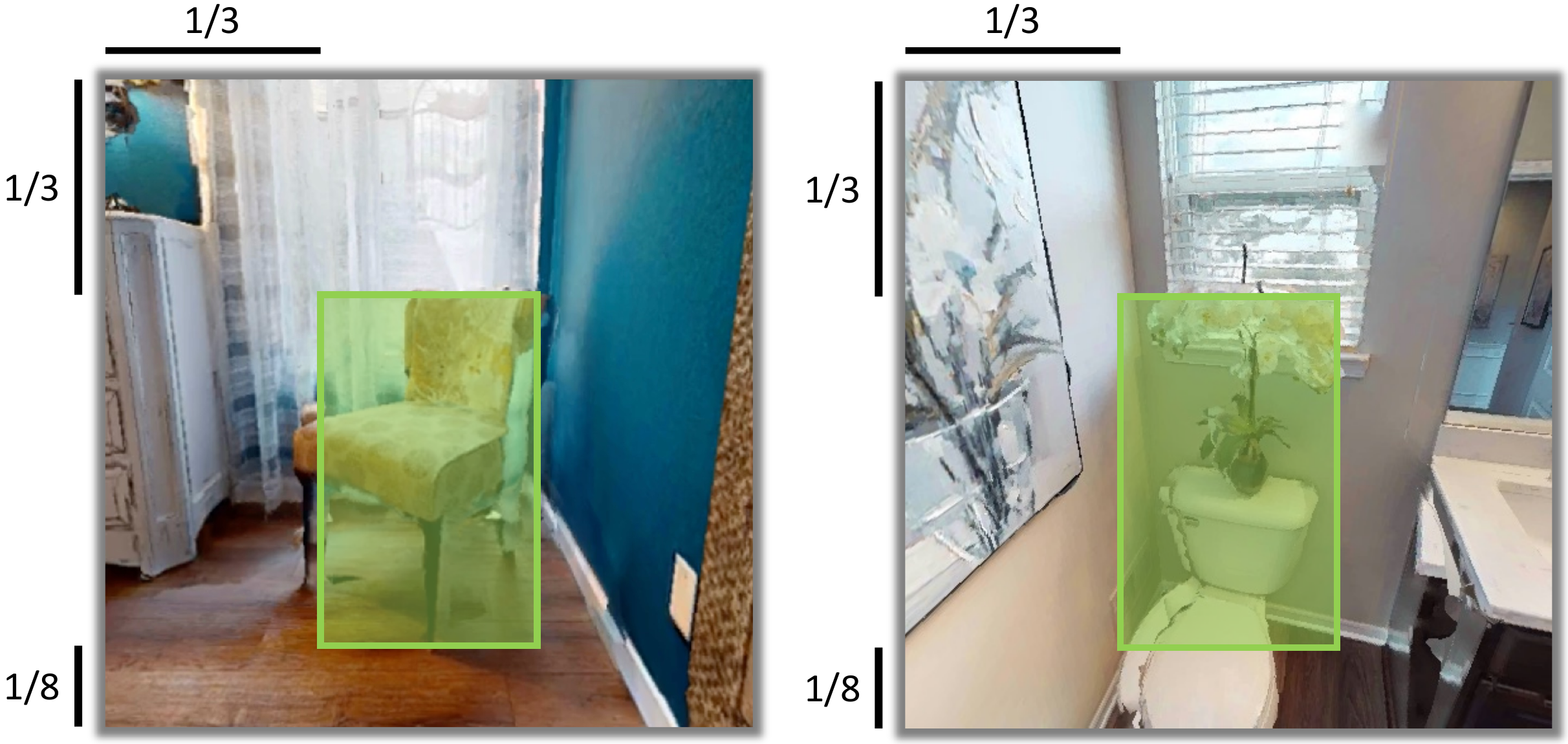}
    \caption{Examples of the static central crop applied to goal images in the Val split. While the crop reasonably frames the chair (left), the plant (right) is much smaller than the crop.}
    \label{fig:central_crop}
\end{figure}

Our primary goal localization method, \textit{Detic-Projected}, involves producing an instance segmentation mask of the goal instance to determine which feature correspondences to project to world coordinates. We ablate this instance mask using the \textit{Crop-Projected} method --- instead of an instance-specific mask, \textit{Crop-Projected} takes a static central crop of the goal image that is the same for all episodes. Matched keypoints lying inside this mask are projected to the goal map channel. We find that taking a crop of the central 1/3 of the goal image extending down to a height of 1/8 to be effective. We visualize two examples of this mask in \figref{fig:central_crop}.

\section{Implementation Details}

We implement our models and vectorized mapping in PyTorch and base our implementation off Home-Robot\footnote{\href{https://github.com/facebookresearch/home-robot}{github.com/facebookresearch/home-robot}} and SemExp ObjectNav\footnote{\href{https://github.com/devendrachaplot/Object-Goal-Navigation}{github.com/devendrachaplot/Object-Goal-Navigation}} repositories. Model variations using ResNet employ a ResNet50 pre-trained on ImageNet-1K. Model variations using CLIP use the \textit{ViT-B/32} weights. We use the pretrained SuperGlue model\footnote{\href{https://github.com/magicleap/SuperGluePretrainedNetwork}{github.com/magicleap/SuperGluePretrainedNetwork}} with indoor-trained weights from ScanNet. For the Detic model, we use the official repository\footnote{\href{https://github.com/facebookresearch/Detic}{github.com/facebookresearch/Detic}} with the default 21K-trained model. We apply a custom vocabulary of $\{$ \textit{chair, sofa, bed, toilet, potted\_plant, tv\_monitor} $\}$. For our real-world experiments, we deploy our system on a Stretch V1 by Hello Robot equipped with the default Intel RealSense D435i RGBD camera.

\section{Real-World Qualitative Example}

\figref{fig:teaser} shows a qualitative example of our agent's performance in the real world. In this example, the agent is operating in Env A (a furnished office space) and is given a picture of a chair. The agent traverses a hallway past a couch (Steps 1, 97), enters the room with the chair (Step 164), re-identifies and localizes the chair (Step 229), and successfully navigates to it (Step 237). This example shows our method's robustness to the domain gap between sim and real -- RGBD observations on Stretch contain real-world noise and poses are estimated using Hector SLAM on ROS\footnote{\href{http://wiki.ros.org/hector_slam}{wiki.ros.org/hector\_slam}}. The image goal is taken from a mobile phone camera and from a pose that affords an intuitive view of the goal instance according to the user that issues the task. With no reliance on robot hardware, this demonstrates the ease of goal specification in InstanceImageNav and the robustness of our method.

\section{Complete Real-World Results}

\tabref{tab:real_world_results} shows the description and results for each episode performed by \texttt{Mod-INN} in the real world. The episode IDs in \tabref{tab:real_world_results} correspond to the IDs of videos on the project page.

\setlength{\tabcolsep}{.3em}
\begin{table}[h]
  \setlength{\aboverulesep}{0pt}
  \setlength{\belowrulesep}{0pt}
  \renewcommand{\arraystretch}{1.15}
  \begin{center}
    \resizebox{0.68\columnwidth}{!}{
      \begin{tabular}{cll s}
        \toprule
        Episode ID & Environment & Goal ID & \textbf{Success?} \\
        \midrule
        \texttt{1} & A & Chair\_0 &  True \\
        \texttt{2} & A & Plant\_0 &  True \\
        \texttt{3} & A & Couch\_0 &  False \\
        \texttt{4} & A & Couch\_0 &  True \\
        \texttt{5} & A & Couch\_0 &  True \\
        \texttt{6} & A & Couch\_0 &  True \\
        \texttt{7} & B & Plant\_1 &  True \\
        \texttt{8} & B & Bed\_0   &  True \\
        \bottomrule
      \end{tabular}
    }
  \end{center}
  \caption{Description of episodes performed by \texttt{Mod-INN} in the real-world with results.}
  \label{tab:real_world_results}
\end{table}

\balance

\end{document}